\RequirePackage{fix-cm}
\documentclass[smallextended]{svjour3}       
\smartqed  
\usepackage{graphicx}
\usepackage[title]{appendix}
\usepackage{enumitem}
\usepackage{cite}
\usepackage{amsmath}
\usepackage{amsfonts}
\usepackage{amssymb}
\usepackage{blindtext}
\usepackage{hyperref}
\usepackage{booktabs}
\usepackage[flushleft]{threeparttable}
\usepackage[linewidth=1pt]{mdframed}
\usepackage{epsfig}

\usepackage{lineno}

\DeclareMathOperator*{\argmin}{arg\,min}
\journalname{Empirical Software Engineering}
\begin{document}

\title{Which Design Decisions in AI-enabled Mobile Applications Contribute to Greener AI?
}


\author{Roger Creus Castanyer         \and
        Silverio Martínez-Fernández \and 
        Xavier Franch
}


\institute{Roger Creus Castanyer \at
              Universitat Politècnica de Catalunya \\
              Barcelona, Spain\\
              \email{roger.creus@estudiantat.upc.edu}\\
              ORCID: 0000-0003-1952-3357
           \and
           Silverio Martínez-Fernández\at
              Universitat Politècnica de Catalunya \\
              Barcelona, Spain\\
              \email{silverio.martinez@upc.edu}
              \\
              ORCID: 0000-0001-9928-133X
            \and
           Xavier Franch \at
              Universitat Politècnica de Catalunya \\
              Barcelona, Spain\\
              \email{xavier.franch@upc.edu}\\
              ORCID: 0000-0001-9733-8830
}

\date{Received: date / Accepted: date}

\maketitle

\begin{abstract}
\textbf{Background}: The usage of complex artificial intelligence (AI) models demands expensive computational resources. While currently available high-performance computing environments can support such complexity, the deployment of AI models in mobile devices, which is an increasing trend, is challenging. Environments with low computational resources imply limitations in the design decisions during the AI-enabled software engineering lifecycle that balance the trade-off between the accuracy and the complexity of the mobile applications.

\noindent \textbf{Objective}: Our objective is to systematically assess the trade-off between accuracy and complexity when deploying complex AI models (e.g. neural networks) to mobile devices in pursuit of greener AI solutions. We aim to cover {\it{(i)}} the impact of the design decisions on the achievement of high-accuracy and low resource-consumption implementations; and {\it{(ii)}} the validation of profiling tools for systematically promoting greener AI.

\noindent \textbf{Method}: We implement neural networks in mobile applications to solve multiple image and text classification problems on a variety of benchmark datasets. We then profile and model the accuracy, storage weight and time of CPU usage of the AI-enabled applications in operation with respect to their design decisions. Finally, we provide an open-source data repository following the EMSE open science practices and containing all the experimentation, analysis and reports in our study.

\noindent \textbf{Results}:    We find that the number of parameters in the AI models makes the time of CPU usage scale exponentially in convolutional neural networks and logarithmically in fully-connected layers. We also see the storage weight scales linearly with the number of parameters, while the accuracy does not. For this reason, we argue that a good practice for practitioners is to start small and only increase the size of the AI models when their accuracy is low. We also find that Residual Networks (ResNets) and Transformers have a higher baseline cost in time of CPU usage than simple convolutional and recurrent neural networks. Finally, we find that the dataset used for experimentation affects both the scaling properties and accuracy of the AI models, hence showing that researchers must study the presented set of design decisions in each specific problem context. 

\noindent \textbf{Conclusions}: We have depicted an underlying and existing relationship between the design of AI models and the performance of the applications that integrate these, and we motivate further work and extensions to better characterise this complex relationship.

\end{abstract}

\section{Introduction}
Artificial Intelligence (AI) plays a key role in the world we live in. AI is about making machines mimic intelligent behaviour. Machine Learning (ML) sets the foundations of AI \cite{mohri2018foundations}. ML works over the most valuable and key source of knowledge that exists: data. In ML, machines take in data and learn patterns that would be difficult for humans to learn. ML is valuable in the sense that the data processing (or data classification, segmentation, representation) capabilities go far beyond than what humans can achieve. ML finds its extension in Deep Learning (DL), which introduces complex neural network (NN) models with the purpose to provide highly performing capabilities in an increasing number of challenging domains \cite{lecun2015deep} (e.g. image-based and language-based contexts). Nowadays, NNs have achieved outstanding results and have outperformed humans in domains as diverse as machine translation, character and handwriting recognition, speech and facial recognition and videogames \cite{tappert1990state, silver2017mastering, oord2016wavenet, mao2017deepart}. Key to the success of NNs is their capability of learning complex data representations in an unsupervised manner. In this way, NNs do not demand expert feature engineering tasks, which are a bottleneck of ML models performance. However, training NNs requires both large amounts of data and high computational resources. These two needs are not a stopper for NNs since we are living the \textit{Big Data, Big Compute, Big Models} revolution \cite{byun2012driving, dowd2010high, brown2020language}. As a consequence, following the exponential evolution of the availability of data and computational power, the tendency of solving complex tasks with complex models is increasing \cite{rasley2020deepspeed, brown2020language, zellers2019defending, devlin2018bert}. This tendency does not take into account resource consumption regulations, and does not promote less data-intensive and lighter models, which are considerations that set the bases of greener AI. For example, recent language models like the GPT-3 have 175 billion parameters \cite{brown2020language}, and benchmarks for image classification have been trained for more than 6 months \cite{hinton2015distilling}. 

For \textit{green AI} we understand AI research and practices that are more environmentally friendly and inclusive \cite{schwartz2019green}.   Green AI refers to the development and deployment of AI systems that minimize their negative impact on the environment and promote sustainable energy and resource consumption \cite{calero2021introduction}. The development of greener AI can be achieved through various means such as optimizing the energy efficiency of hardware used for AI processing, reducing the carbon footprint of data centers that host AI systems, and developing algorithms that promote sustainable practices. In contrast, \textit{red AI} refers to AI research that seeks to improve accuracy (or related measures) through the use of massive computational power while disregarding the environmental cost \cite{schwartz2019green}. Ensuring efficient implementations is key to achieve greener AI. Some of the key practices of green AI are to measure and lower the environmental impact of AI \cite{lacoste2019quantifying}, regulate the resource consumption of the AI-enabled applications \cite{8805694} and balance the effectiveness and efficiency of the AI models \cite{tan2019efficientnet, sanh2019distilbert}. Central to these practices is the capability of measuring and monitoring the efficiency-related metrics of AI-enabled applications in operation, which allows developers to gain awareness of the performance of AI-enabled applications with respect to their complexity.

Deploying AI models to operate in environments with low computational resources constrains the design decisions in AI-enabled applications\footnote{note that in this work, whenever we relate to AI models or AI components we mean NNs, since these are the ones that we study, develop and test in this paper}. One of such low computational resources environments are mobile applications. With the growth of the mobile applications market, approaches on the integration of AI models in lightweight devices are becoming popular given their ability to provide great services to the user \cite{deployment-DL}. In the context of AI-enabled mobile applications, reducing energy consumption, time of CPU usage (i.e. in CPU or GPU) and storage weight is fundamental for deploying a user-friendly experience \cite{citeKey}. However, AI practitioners aim at providing high-accuracy results, which might involve deploying computationally demanding AI models, so balancing the efficiency and effectiveness of AI-enabled mobile applications is key to pursuing greener AI. In this paper, we approach green AI solutions by developing AI-enabled mobile applications whose focus is to maximize their accuracy while keeping their complexity under control.

In this work we use mobile devices as environments that support the deployment of AI models. Few frameworks are available that enable the end-to-end implementation of AI models into mobile applications, and such frameworks have a strong influence in the overall applications performance \cite{empirical-frameworks}. For this reason, there exists the need of a systematic approach for measuring the performance of an AI-enabled mobile application that takes into account the accuracy and complexity-related metrics.

With this, we aim to model and investigate the contribution of design decisions during the AI-enabled software engineering lifecycle to the overall mobile applications performance in terms of accuracy and complexity. Our objective is to report statistically significant relationships between the design decisions and their impact in order to build the bases for predictive tools that empower practitioners with consciousness of the performance of their AI-enabled applications. 

This article is structured as follows. Section 2 introduces the Related Work; Section 3 describes the experimental design in detail (i.e. variables, hypotheses, research questions); Section 4 presents the experiment's results and findings; in Section 5 we discuss the implications of our work for AI engineering research and industry; in Section 6 we identify the threats to validity; and finally in Section 7 we present our conclusions and motivate future work on the topic.  

\section{Related work}
Engineering AI-enabled applications that operate in edge devices is challenging \cite{castanyer2021integration, large-scale-AI, deployment-DL, martinez2021}. By edge devices, we understand any device that can host an AI-enabled application so that the AI models run in local machines (e.g. PCs, laptops, mobile phones or even IoT devices) and not in the cloud. Independently of the platform where the AI-enabled software is deployed, the development lifecycle for AI-enabled software, which is significantly different from the traditional software one \cite{9238323}, can be split into 4 phases. First, the \textit{Data Management} which consists of the collection and processing of raw data. Second, the \textit{Modelling} of the AI components which consists of adjusting different models to obtain the best-performing possible solution. Third, the \textit{Development} of the environment where the models will be deployed and the integration of the AI models in it. Fourth, the \textit{System Operation} of the AI-enabled applications. This lifecycle introduces concerning challenges that are not usual in traditional software engineering \cite{quality-AI}. Data centricity is what increases the risk in many quality attributes (e.g. adaptability, scalability, explainability, privacy) \cite{large-scale-AI}. As improving the sustainability of the aforementioned AI-based software engineering lifecycle has never been a key requirement historically, now there exists a high potential for improving the efficiency of the lifecycle and the related development tools \cite{Calero2017Chapter1I}.

\subsection{Research to build Greener AI systems}

In pursuit of greener AI, several works have put the focus on improving energy efficiency \cite{verdecchia2023systematic}.   In the context of traditional mobile applications (without AI components), it has been shown that different works deal with energy efficiency using optimizations of different nature \cite{bao2016android}. For example, leveraging automatic refactoring tools to fix specific code smells in mobile applications has been shown to improve energy efficiency in practice \cite{cruz2018using, banerjee2016automated,xu2023energy}. Another approach to study and optimize energy efficiency consists of monitoring  the application's performance using profilers \cite{7884613}. Monitoring energy consumption, which has been studied from the software (e.g. energy profilers) and hardware (e.g. power monitors) viewpoints, is not trivial \cite{7884613, cruz2019catalog}. In the majority of cases, when measuring energy consumption, one cannot isolate the software from the system where it is running. The latter can be achieved by making sure that no unnecessary applications or processes are running on the device\footnote{\url{https://luiscruz.github.io/2021/07/20/measuring-energy.html}}. Some examples of energy profilers are the Intel Power Gadget and Intel PowerLog, which provide detailed energy consumption profiled data (in Joules) plus CPU utilisation (in \%) and power (in Watt). However, note that these tools are meant to run on a computer machine and not on a mobile device. For this reason, in this work we adopt a technology stack that allows practitioners to both deploy AI-enabled applications into mobile devices and monitor the CPU utilization directly in the edge devices using profilers.

More recently, researchers have found that software energy consumption can be modeled (and thus, estimated) accurately \cite{7884613, chowdhury2019greenscaler}. The CPU usage information has been demonstrated to be very valuable for achieving accurate energy consumption estimations \cite{chowdhury2019greenscaler, georgiou2022green}. In this way, the CPU usage, which is popular for being easily monitored, is a great representative of the energy cost of an application. In light of the results learned from these works, we include the time of CPU usage as one of the dependent variables of the study and define a set of design decisions and provide a novel interpretation of the relationship between these.

In the context of AI-enabled mobile applications, studies on the frameworks for developing the AI components have been proven to cause differences in the overall performance of the applications. Concretely, two popular frameworks which are Pytorch and Tensorflow have been found to carry differences in the accuracy \cite{empirical-frameworks} and also in green characteristics like the energy consumption \cite{georgiou2022green}. In this work, we also use PyTorch and Tensorflow to implement the AI models.
 
Green AI solutions have also been studied from model-centric \cite{xu2023energy} and data-centric viewpoints. In this context, it has been observed empirically that optimizations on the datasets can be performed to reduce the energy consumption of the AI models without significant losses of accuracy \cite{verdecchia2022data}

Compared to the aforementioned works, we adopt a hybrid approach between the model-based and data-centric approaches for constraining resource consumption. In this context, our study is novel in exploring design decisions for both the AI models and the datasets used and how these affect the greenability (i.e., energy efficiency) of the AI-enabled mobile applications.\\

\section{Experimental Design}
\subsection{Research Objectives}
We define our goal and Research Question (RQ) following the GQM guidelines \cite{GQM}.

Overall, our goal is to
analyse \textit{the design decisions of AI-enabled applications}
with the purpose to \textit{assess their impact} 
with respect to \textit{green characteristics} from the point of view of \textit{a developer} 
in the context of \textit{mobile applications with neural networks for image and language classification}. 

We determine the units of analysis and formulate the RQs as follows. As aforementioned above, we work towards deploying green AI-based mobile applications by putting the focus on optimising their performance, defined as the trade-off between accuracy and complexity. We aim to {\it{(i)}} evaluate and quantify the implications of a set of key design decisions that we identify along the AI-enabled software lifecycle with respect to the overall application performance; and {\it{(ii)}} leverage profiling tools for measuring the performance and support the decision-making in the design of AI-enabled applications. 

We establish research objectives, variables and hypotheses from which we perform an empirical study. In particular, we study the systematic variation of design decisions to measure their influence on the accuracy and complexity-related metrics in an experimental set-up. For this purpose, we implement and monitor AI-enabled applications in the image and language classification domains. We profile the performance of the applications by taking into account their time of CPU usage, storage weight and the accuracy in operation. Finally, we define linear regression equations for measuring the implications of the design decisions to the AI-enabled mobile applications performance. In this sense, our research objectives consist on finding statistically significant relations between dependent and independent variables. As no statistical method can show causation, we aim to provide a well-suited regression analysis for minimizing the error in the causal interpretation of our models \cite{miles1994qualitative}. 

With all this, we identify two design decisions that are related to the AI models, which are the number of parameters and the architecture type, and also one related to the data which is the dataset used for training the AI models. The number of parameters represents the total sum of weights and biases that operate inside the AI models (i.e. the NNs). The architecture type controls the degree of complexity introduced in the models and correlates with the amount of computational power provided. The dataset used for training conditions the quality and quantity of the data that feeds the models, hence affecting the models' performance (e.g. in terms of accuracy).

In this way, whenever we refer to the design decisions in this paper, we refer to this set of three variables that we have identified. Then, as stated above, we characterise the performance of the AI-enabled mobile application in terms of the accuracy and complexity trade-off and for that we measure the time of CPU usage, storage weight and the accuracy of the applications in operation. For these reasons, we establish the RQs as follows:  

\begin{itemize} 
\item{RQ1 - What is the impact of the design decisions on the performance of AI-enabled applications in terms of accuracy and complexity?}
    \begin{itemize}[label=$\bullet$]
    \item{RQ1.1 - How is the time of CPU usage of the AI models affected by their number of parameters, architecture type and dataset used?}
    \item{RQ1.2 - How is the storage weight of the AI models affected by their number of parameters, architecture type and dataset used?}
    \item{RQ1.3 - How is the accuracy of the AI models affected by their number of parameters, architecture type and dataset used?}
\end{itemize}
\end{itemize}

With \textit{RQ1} we aim to validate our criterion for reasoning about high-performance implementations of AI models in AI-enabled software applications (i.e., which provide high-accuracy results while keeping complexity at a low level). We aim to provide a quantitative analysis of the importance and relation between design variables and performance metrics in terms of accuracy and complexity. Concretely, we aim to deploy a quantitative evaluation of the performance of the AI-enabled applications in operation. With this, our goal is to delineate the differences in performance caused by the AI models configuration (i.e. architecture type, number of parameters) and dataset used, and to validate them across the vision and language classification domains.

As we frame our study to the statistical modelling of the green characteristics from a set of design decisions, the validity of our hypotheses aims to be independent of whether we prove an existent relationship between our objects of study or we discard it. In this way, we deny any confirmation bias related to prior beliefs of ours on the possibly existing relations between the variables in the study.  Moreover, we perform repeated experimentation, so the results in the profiled metrics and accuracy measurements are averaged across at least 25 examples which can be found in our public repository\footnote{\url{https://github.com/roger-creus/Which-Design-Decisions-in-AI-enabled-MobileApplications-Contribute-to-Greener-AI/tree/main/data/accuracy\%20tests}}. Finally, we discuss and perform the statistical analysis of the results in consensus between all the authors to generalize our conclusions and remove personal interpretation biases.

\begin{figure}
    \centering
    \includegraphics[angle=90,origin=c, scale=0.5]{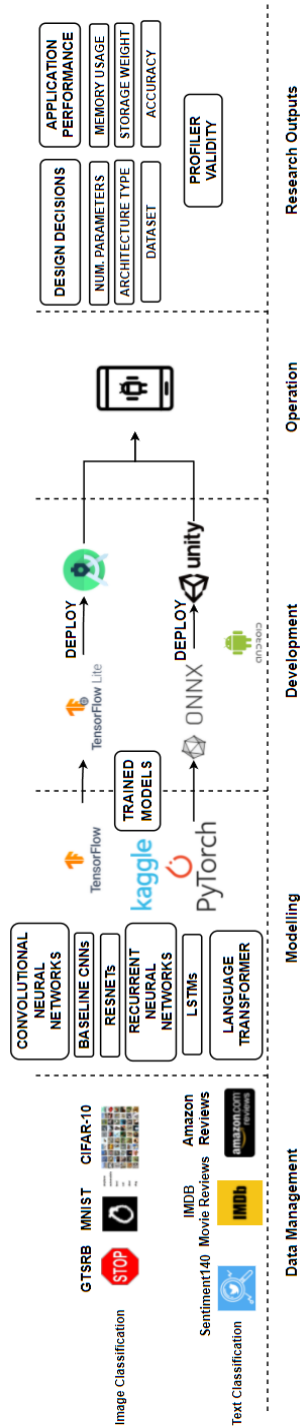}
    \caption{Schema of our empirical study in the end-to-end software engineering lifecycle for AI-enabled applications. In the {\it{(i)}} Data Management phase we obtain and preprocess the data sources; in the {\it{(ii)}} Modelling phase we train the AI models for solving the defined tasks   using the free NVIDIA TESLA P100 GPU hours offered weekly by Kaggle}; in the  {\it{(iii)}} Development phase we convert and deploy the trained models in the applications that we build; in the  {\it{(iv)}} Operation phase we make use of the applications and monitor their performance within the operation settings by means of profilers. Finally, we provide answers to the RQs.
    \label{schema}
\end{figure}

The development schema of the empirical study can be seen in Figure \ref{schema}.

\newpage

\subsection{Variables}\label{variables}

In the following subsections, we respectively report the independent, dependent, and confounding variables of our experimental design.

\subsubsection{Independent Variables}
In this study, we define multiple independent (i.e. explanatory) variables in order to evaluate the contribution of the design decisions to the overall applications performance. 

Regarding the data management phase of the software engineering lifecycle for AI-enabled applications, we experiment with a set of different datasets for each of the vision and language domains. In this way, we define the dataset (D) nominal variable indicating which dataset is used for training the AI models. Implicitly, each dataset provides different quantity and quality of data and this certainly affects the performance of the models. Concretely, the \textit{dataset} takes values in a subset of the most popular benchmark datasets in the vision and language domains: the German Traffic Sign Recognition Benchmark (GTSRB) dataset, MNIST, and CIFAR for solving image classification, and Sentiment140, IMDB Movie Reviews and Amazon Reviews for text classification (see Section \ref{dataset-section} for a detailed description of the datasets). We chose these datasets since they are well-known benchmarks which can be solved with NNs and at the same time carry significant differences in terms of the quality and quantity of the data (e.g. CIFAR contains 32x32 RGB images and MNIST contains 28x28 grayscale images).   The presented datasets have many thousands of monthly downloads in HuggingFace \footnote{\url{https://huggingface.co/datasets?sort=downloads}}, and as mentioned above, we believe that all together they form a rich population of diverse datasets in terms of quality in the quantity of data.

Regarding the modelling phase of the AI components, we experiment with {\it{(i)}} different number of parameters (P), which is a numerical variable indicating the complexity of the AI models; and {\it{(ii)}} different architecture types (AT), encoded as a nominal variable characterizing different model architectures (in a subset of the SOTA architectures formed by the LSTM (Long Short-Term Memory), GRU (Gated Recurrent Unit), Transformer, CNN (Convolutional Neural Netwoek), ResNet (Residual Networks) and MobileNets).    The idea behind studying the \textit{number of parameters} is to measure how scaling up the AI models in terms of computational power helps with achieving better accuracy and how it impacts the performance of the applications. Nowadays, many successes in the AI literature rely on very deep AI models trained in computational environments that are inaccessible to the vast majority of researchers. For this reason, we aim to understand the role of the scale of the AI models in the overall AI-enabled applications to help practitioners make the best use of it in any environment. With the \textit{architecture type}, we aim to study how more complex architectures in both the vision and language domains (e.g. ResNets, Transformers) help achieving better accuracy or performance in the AI-enabled applications. These architectures are based on the simple convolutional and recurrent neural networks (that we also include in our study) but usually involve some novel operations (e.g. residual layers in ResNets, attention mechanism in Transformers) that are worth studying.

\subsubsection{Dependent Variables} \label{dependent-variables}

We evaluate the contribution of the design decisions with respect to the overall performance of the built applications in operation, which we study as a balance between the accuracy and the complexity. To define the performance of AI-enabled applications, we consider three dependent variables, the first two referring to the concept of complexity and the latter representing the accuracy. 

\begin{itemize}
    \item \textit{Time of CPU usage} (M), which quantifies the resource consumption that the applications require to run (in ms). 
    \item \textit{Storage weight} (S), which quantifies the cost of deploying the AI models to the applications and storing them (in MB).
    \item \textit{Accuracy} (A), which measures how capable is the application (and the AI model) of providing high-accuracy and satisfactory results in operation   (\% of correctly classified inputs across an evaluation dataset \footnote{\url{https://github.com/roger-creus/Which-Design-Decisions-in-AI-enabled-MobileApplications-Contribute-to-Greener-AI/tree/main/data/accuracy\%20tests}}).
\end{itemize}

Importantly, in our experiments we only report the time of CPU usage and do not study the impact of using GPUs or other modern dedicated AI-processing cores (e.g. Neural Processing Units). As can be seen later in the results section (e.g. Figure \ref{unity_profiler}), all the AI model usage falls under the CPU module.

Regarding accuracy, we plan to measure the accuracy of the AI models when operating in the AI-enabled applications during the system operation phase of the AI-enabled software lifecycle. That is, we test the accuracy of the models with data outside the context of the training datasets to test the models’ generalization capabilities. Concretely, we craft a test dataset for each domain (vision and language) containing data from the operation environment with a sufficiently large number of examples to test the applications systematically and provide statistically significant conclusions on the accuracy achieved. For the image domain we plan to report the top-3 accuracy (which considers a model output to be accurate if it contains the actual class in the top-3 predicted classes by the model when ranked by confidence), as we consider that the top-1 accuracy might be too demanding for large multi-class classification problems). For the language (i.e. text) domain we report the top-1 accuracy as text classification is generally a problem with a smaller number of classes.

\subsubsection{Confounding Variables \& others}
We identify a confounding factor related to the experience and knowledge of the developer who builds the AI models and applications, since it conditions the overall performance of the AI-enabled applications in terms of their accuracy and complexity. A developer is responsible for the implementation of both multiple sophisticated AI models and mobile applications that integrate them. Moreover, he is also in charge of operating with the AI-enabled mobile applications, profiling and analyzing the applications’ performance during the system operation phase. We consider that the more experience and knowledge the developer has, the less overhead in the AI-enabled applications. Ideally, as we aim to profile the performance of AI-enabled applications, we want to  account only for the performance changes produced by our objects of study and not by deficient implementations. For mitigating the latter we provide as simple implementations as possible by only implementing the required functionalities to the application and by following official tutorials for the development of these.

\begin{table*}[h!]
\centering
\caption{\label{tab:variables}The variables of the study}
    \begin{tabular}{p{0.13\textwidth}p{0.18\textwidth}p{0.24\textwidth}p{0.09\textwidth}p{0.23\textwidth}}
        \hline
        \textbf{Class} & \textbf{Name} and \textbf{Abbreviation} & \textbf{Description} & \textbf{Scale} & \textbf{Operationalization}  \\
        \hline
         Independent & Number of Parameters (P) & The number of trainable parameters of the NNs & numerical & See Sections \ref{variables} and \ref{hypotheses}\\
          & Architecture Type (AT) & The architecture that defines the NNs functionality & nominal & See Sections \ref{variables} and \ref{hypotheses} \\
          & Dataset (D) & The dataset used for training the AI models & nominal & See Section \ref{dataset-section} \\
         \hline
         Dependent & Time of CPU usage (M) & The time that the AI-enabled applications need to answer a query & numerical & Profiled \\
         &  Storage Weight (S) &The cost of storing the AI models in the edge devices & numerical & Retrieved from the built applications \\
         & Accuracy (A) & The precision of the AI models when answering queries in the edge devices & numerical & Measured in the system operation phase (see Section \ref{dependent-variables}) \\
         \hline
         Confounding \& others & Developer experience & The ability to implement high-quality software applications & - & - \\
         & AI modelling framework & Technology used for modelling the AI components & - & Experiment with Pytorch and Tensorflow \\
         & Export AI components & Technology used for exporting the AI components & - & Experiment with ONNX and Tensorflow Lite \\
         & AI-enabled development framework & Technology used for developing the AI-enabled applications & - & Experiment with Unity3D and Android Studio \\
         \hline
    \end{tabular}
\end{table*}

We also consider the technology stack used for deploying our experimentation a factor that conditions the results obtained. For that we experience with two different frameworks for modelling the AI components which are Pytorch and Tensorflow; we use two different technologies for exporting the trained AI models which are ONNX and Tensorflow Lite; and we also use two different frameworks for developing the AI-enabled mobile applications: Unity3d and Android studio. With this pipeline, we provide experiences with two different technology stacks that enable practitioners to deploy AI models in mobile applications. In this way, we can assess if there exists noticeable performance differences between the applications developed with different technologies and we can also suggest the most suitable tools and practices for developers willing to integrate and profile AI models in mobile applications.  

\subsection{Hypotheses and Analysis Operationalization} \label{hypotheses}
In this study we hypothesize that there is an underlying relation between the design decisions of AI-enabled software and their consequences and we aim to characterize it. Concretely, we study the accuracy (A) and the complexity (M, \textit{storage weight}) as a model of the design decisions (\textit{number of parameters}, \textit{architecture type}, \textit{dataset}), with separate hypothesis for the vision and language domains. The models that we consider are linear models between the design decisions (i.e. independent, explanatory variables) and the accuracy and complexity-related metrics (i.e. dependent, response variables). When fitting linear models we provide the following capabilities:

\begin{itemize}
    \item Determining the goodness of fit by quantifying the amount of variability of the dependent variables that is captured (i.e. explained) by the set of independent variables.
    \item Using the fit as a predictive model for regressing accuracy and complexity responses for untested experimental conditions.
    \item Quantifying the strength of the relationship between the dependent and independent variables, and determining whether some independent variables may have no linear relationship with the responses at all, or identifying which subsets of independent variables may contain redundant information about the responses.
    \item Providing directional analysis on what independent variables contribute more to the target dependent variables.
    \item Determining whether the interactions between pairs of dependent variables significantly contribute to the independent variables.
\end{itemize}

With all this, we define the models as follows:

\begin{flalign}
    &M = \beta_{P_1} P + \beta_{AT_1} AT + \beta_{D_1} D + \epsilon_1 \label{MModel} \\
    &S = \beta_{P_2} P + \beta_{AT_2} AT + \beta_{D_2} D + \epsilon_2 \label{SWModel} \\ 
    &A =  \beta_{P_3} P + \beta_{AT_3} AT + \beta_{D_3} D + \epsilon_3 \label{AModel}\\
    \nonumber
\end{flalign}

With equations \ref{MModel}, \ref{SWModel}, \ref{AModel} we define multiple linear regressions to explain the variability of the time of CPU usage, storage weight and accuracy with respect to the design decisions separately (1: M; 2: S; 3: A) \cite{lai1979strong}. For example, with $\beta_{P_1}$ we study the impact of the number of parameters only to the time of CPU usage. Ideally, we want all the variables $\epsilon_i$ to be of low magnitude, as it represents the unexplained variance of the dependent variables.

We fit each of the three multiple linear models with the least-squares estimation technique. In particular, we fit the coefficients $\beta_i$ (for each of the design decisions) to minimize the squared error between model predictions and observed values (belonging to the data collected in our experimentation):

\begin{equation}
    \boldsymbol{\hat{\beta_i}} = \argmin_{\boldsymbol{\beta_i}} \sum_{n=1}^{N} (\boldsymbol{\beta_i} * \textbf{x}_n - \textbf{y}_{n_i})^2
\end{equation}
where $\boldsymbol{\hat{\beta_i}}$ is the vector of fitted coefficients of the \textit{i-th} model, $N$ is the number of experiments that we perform, $\textbf{x}_n$ are the conditions of the \textit{n-th} experiment, and $\textbf{y}_{n_i}$ is the observed response variable $i$ that we profile in the \textit{n-th} experiment (following the association of $i=$ 1: M; 2: S; 3: A). 

Then, we first test whether equations \ref{MModel}, \ref{SWModel}, \ref{AModel} are suitable models of the aforementioned relations. We do so with an F-test for each of the three multiple linear regressions. We define the hypotheses of the \textit{i-th} F-test as follows, for $i=$ 1: M; 2: \textit{storage weight}; 3: \textit{accuracy}.

\begin{flalign}
  \mathcal{H}_{0_i} &: \beta_{P_i} = \beta_{AT_i} = \beta_{D_i} = 0 & & \textnormal{(Null Hypothesis)}\nonumber\\
\mathcal{H}_{1_i} &:  \exists \beta \in \{\beta_{P_i}, \beta_{AT_i}, \beta_{D_i}\} : \beta \neq 0 & & \textnormal{(Alternative Hypothesis)}\nonumber
\end{flalign}

In this sense, we first test whether there is a linear relation between each of the independent variables and the dependent ones. In the F-test, we compare the fitted model sum of squares against the residual sum of squares, and calculate the value of the F-statistic to derive the p-value associated to a level of confidence and to reject/support the null hypothesis \cite{pope1972use}. 

Furthermore, we can evaluate how valuable the fitted models are with the adjusted $R^2$ statistic \cite{miles2014r}. In our multiple linear regression models, the $R^2$ is the square of the multiple correlation coefficient, which measures how well a response variable can be predicted with a linear function of a set of independent variables.

We also formulate an hypothesis for each explanatory variable in each of the models of the independent variables ($\mathcal{H}_{P_i}$, $\mathcal{H}_{AT_i}$, $\mathcal{H}_{D_i}$) that concerns the statistical significance of each design decision in each of the specific models. Hence, we have three hypotheses for each model which sums to a total of another 9 hypotheses. With these, we aim to separately answer the research questions RQ1.1, RQ1.2 and RQ1.3. Each of these hypotheses can be tested with an adjusted t-test, which is an inferential statistic used to determine if there is a significant difference between the means of two groups \cite{student1908probable}. Concretely, we aim to determine whether the coefficients encoding the contribution of each of the design decisions to the accuracy and complexity-related metrics are significantly different from zero. If the fitted coefficient corresponding to a design decision, e.g. \textit{number of parameters} of the AI models, happens to be significantly different from zero and take a value \textit{y} in a specific model $i$, we can state that if we fix all the other non-zero coefficients of model $i$, then for each change of 1 unit in the \textit{number of parameters}, the response variable of model $i$ changes \textit{y} units.

\begin{align*}
  \mathcal{H}_{{P_0}_i} &: \beta_{P_i} = 0 & & \textnormal{(Null Hypothesis)} \\
  \mathcal{H}_{{P_1}_i} &: \beta_{P_i} \neq 0 & & \textnormal{(Alternative Hypothesis)} \\
  \mathcal{H}_{{AT_0}_i} &: \beta_{AT_i} = 0 & & \textnormal{(Null Hypothesis)} \\
  \mathcal{H}_{{AT_1}_i} &: \beta_{AT_i} \neq 0 & & \textnormal{(Alternative Hypothesis)} \\
  \mathcal{H}_{{D_0}_i} &: \beta_{D_i} = 0 & & \textnormal{(Null Hypothesis)} \\
  \mathcal{H}_{{D_1}_i} &: \beta_{D_i} \neq 0 & & \textnormal{(Alternative Hypothesis)}
\end{align*}

These hypotheses test whether a specific independent variable (e.g. P) is significantly adding to the variability of the \textit{i-th} variables of study given that the \textit{i-th} model also considers other explanatory variables (e.g. \textit{architecture type}, \textit{dataset}). 

With all this, we aim to {\it{(i)}} determine whether the accuracy and complexity of an AI-enabled application can be modelled by taking into account the design decisions; {\it{(ii)}} investigate the contribution of each design decision, analyse their interactions, and determine the most impactful design decisions on the overall performance of the applications. 

For this reason, we explore a complete set of representative combinations of experimental settings. Concretely, we fix all the design decisions and iterate variations on a single one (and we do this for each of the design decisions) to consistently quantify the contribution of each of the design decisions.

\subsection{Execution of the empirical study} \label{execution}
In this section we explain how we have followed the empirical study design depicted above in Figure \ref{schema} and defined in our registered report \cite{castanyer2021design}. Following the open science initiative of the EMSE journal \cite{mendez2019open}, we share a replication package available at \url{https://doi.org/10.5281/zenodo.6523801}. The replication package enables to reproduce the experiments described in Section \ref{modelling-section}. Concretely, it consists of the source code for {\it{(i)}} developing the AI models (in Python); {\it{(ii)}} the source code of the AI-enabled mobile applications (in C\#); {\it{(iii)}} the source code for the statistical analysis (in R); and {\it{(iv)}} all the data needed for the aforementioned tasks. Deviations are also reported.    With {\it{(i)}} we implement and train the AI models; with {\it{(ii)}} we develop the AI-enabled applications, build and deploy them to the mobile devices, and use the profilers to measure their performance. During operation, we also use the applications to query them with multiple inputs and measure their average accuracy and performance; and with {\it{(iii)}} we conduct the statistical analysis on the data generated during the repeated experiments in operation.

\subsubsection{Data management: datasets} \label{dataset-section}
There exists a wide variety of NN models (e.g. recurrent, convolutional, Transformer-based) for tackling different problems. To provide generalist conclusions about AI-enabled applications in the vision and language domains, we aim to study a complete set of representative NN architectures. Also, we want to test the aforementioned architecture types in multiple benchmark datasets to provide a model and data-centric context. For this reason, we make use of the German Traffic Sign Recognition Benchmark (GTSRB) \cite{Benchmarking-tsr}, MNIST \cite{deng2012mnist} and Cifar-10 \cite{krizhevsky2009learning} datasets for solving image classification, and the Sentiment140 \cite{go2009twitter}, Amazon Reviews \cite{ni2019justifying} and IMDB Movie Reviews \cite{maas-EtAl:2011:ACL-HLT2011} for solving text classification.

For image classification, the GTSRB dataset is the output of an attempt to benchmark traffic sign recognition and it contains 51,840 images representing 43 unique traffic sign classes. MNIST is a benchmark dataset of handwritten digits (i.e. contains 10 unique classes belonging to the digits) containing 60,000 training examples, and a test set of 10,000 examples. Cifar-10 is a generic dataset for image classification containing 60,000 32x32 colour images in 10 classes, with 6,000 images per class.

For text classification, Sentiment140 contains 1,600,000 tweets annotated with either \textit{positive} or \textit{negative} labels, which encode the sentiment of the tweets. Then, the Amazon Reviews for Sentiment Analysis\footnote{\url{https://www.kaggle.com/bittlingmayer/amazonreviews}} contains 4 million reviews also tagged either as positive or negative, and the IMDB Movie reviews\footnote{\url{https://www.kaggle.com/lakshmi25npathi/imdb-dataset-of-50k-movie-reviews}} dataset contains 50,000 reviews also for binary sentiment analysis.

\subsubsection{Modelling}  \label{modelling-section}

For image-based AI-enabled applications, we develop multiple Convolutional and ResNet-based neural networks, and for text-based ones we develop multiple LSTMs (Long Short-Term Memory), GRUs (Gated Recurrent Units) and Transformers. Then, we integrate them in Android mobile applications using Unity3D and Android Studio. For the statistical analysis of the accuracy and complexity relations we proceed as follows. First, we train the AI models and annotate their number of parameters (P), their architecture type (AT), their storage weight (S) (before being integrated in the mobile applications) and the dataset (D) used for training them. Secondly, we integrate the AI models in the mobile applications and profile the time of CPU usage (M) which we relate to the time consumption for answering a query and measure the accuracy in the AI-enabled system operation (A). Then, we export the profiled results and post-process them to aggregate all the operations that happen in the models (both in initialization and inference). Finally, we merge all the aforementioned information and craft an analysis dataset for the image and text domains (see the analysis datasets in the replication package provided).

\begin{figure}[h!]
    \centering
    \includegraphics[width=\linewidth]{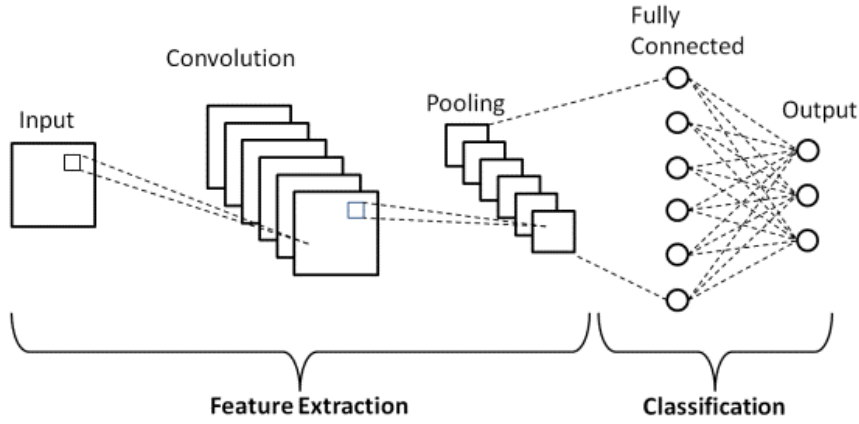}
    \caption{Example architecture of a CNN. }
    \label{cnn-architecture}
\end{figure}

Regarding the execution of our experiments, we perform some additional steps that we describe in the following. For image-based AI models, we add an extra categorical annotation to the Convolutional models (and not to the ResNet-based ones) indicating in which part of the model architecture the change in parameters (i.e. both addition or subtraction of weights and biases) happens. We find the need of making this distinction because we observe that parameters in the convolutional layers and in the fully connected layers behave differently, and hence can be studied separately. Still, note that the variable \textit{number of parameters} indicates the total number of parameters and the extension just indicates which part of the model (convolutional, fully connected) contains the majority of the parameters. Hence we consider mainly convolutional CNNs (C-CNNs) and mainly fully-connected CNNs (FC-CNNs). For the ResNet-based models, we consider the five standard models which are the ResNet18, ResNet34, Resnet50, Resnet101 and Resnet 152. These have conceptually the same architecture and just vary in their complexity. We do not make a distinction between C and FC extensions for the ResNet-based models as they scale in a balanced way within these five models. In Figure \ref{cnn-architecture} we show an instance of a CNN architecture. Whenerver we relate to C-CNNs we mean that we increase the number of filters and layers in the \textit{Feature Extraction} part, and when we relate to FC-CNNs we mean that we increase the number of neurons and layers in the \textit{Classification} part.

For text-based AI models, we also encounter relevant differences in the behaviours of what we call the \textit{embedding} and \textit{network} parameters. The former are stored in the embedding layer, which is a look-up table, and are in charge of learning representations for the word tokens. There are \textit{ tokens * embedding dimension} embedding parameters in a text model, where \textit{tokens} are the number of unique words in the dataset and \textit{embedding dimension} is a hyperparameter to set. Then, the network parameters are in the neural network (e.g. in the recurrent and fully connected layers of the LSTM and in the attention and fully connected layers in the Transformers) and are in charge of learning the semantic relations between the words. Hence, we make the difference between EMB-Transformers, EMB-LSTMs, NN-Transformers and NN-LSTMs indicating which layers in the models dominates in numbers of parameters (in the same way we do for FC-CNNs and C-CNNs) and study them separately.

\subsubsection{Development and Operation}

As aforementioned in Section \ref{execution}, during operation we query the AI-enabled mobile applications with evaluation sets containing 25 input examples each, which are all available in the replication package. We then measure the average accuracy and performance using the profilers and export the results for statistical analysis.

For our experiments, we have used a Xiaomi Poco X3 NFC mobile device  to install and run the AI-enabled applications, which contains an Octa-core (2x2.3 GHz Kryo 470 Gold \& 6x1.8 GHz Kryo 470 Silver) CPU.

During the experiments, we found several incompatibilities between the technologies that we use which block some paths defined in the scope of our study. First, we experience that the MobileNets architecture for images can be deployed in Android Studio but not in Unity due to an unsupported operation in the model \cite{castanyer2021integration}. Secondly, the GRU architecture cannot be imported neither in Android Studio nor in Unity, since it is a quite unpopular architecture compared to LSTMs and Transformers.

Finally, we have experienced that the profiler tools integrated in Android Studio and Unity work in a very different way and both make it quite difficult to export the profiled results into a file format that can be analysed in a simple way (i.e. a CSV). As we were incapable of doing so in Android Studio, we stick to an external package of Unity named \textit{Profiler analyzer} which allows us to export the original outputs of the profiler (a binary file) into a CSV table that we can later analyze using the R programming language. Hence, the following results section is based on the models developed with  Pytorch, exported with ONNX, and operated and profiled in the mobile applications built with Unity. The experimentation to carry out with this technology stack is available in the replication package provided.

\section{Experiments' Results}
In this section, we answer our RQs by testing our hypotheses (see Section \ref{hypotheses}). In the following, we provide details on the experimentation and statistical analysis when operating with the AI-enabled mobile applications in each of the two domains of image and text. Concretely, we define a section for each RQ, and a subsection for each sub-RQ1 corresponding to the analysis of each dependent variable. In the sub-RQ1s we analyse the contribution of all the independent variables of study (i.e. \textit{number of parameters}, \textit{architecture type} and \textit{dataset}) to each of the dependent ones (M, \textit{storage weight} and A) for both domains of image and text. Note that we do this for all the datasets at the same time.   The tables of results that gather all the evidence collected in the experiments are available in our public repository \footnote{\url{https://github.com/roger-creus/Which-Design-Decisions-in-AI-enabled-MobileApplications-Contribute-to-Greener-AI/tree/main/data/results}}.

\subsection{RQ1.1: How is the time of CPU usage of the AI models affected by their number of parameters, architecture type and dataset used?}

We first answer RQ1.1 which regards the relation between the the \textit{time of CPU usage} and our independent variables of study. For that, we study the relation between the \textit{number of parameters}, \textit{architecture type}, the \textit{dataset} and the \textit{time of CPU usage} in the two domains of image and text.

\subsubsection{Image-domain}
\begin{figure}[h!]
    \centering
    \includegraphics[width=\linewidth]{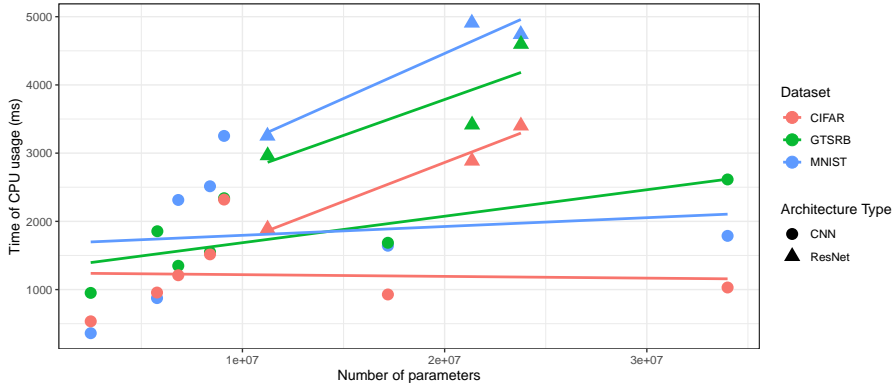}
    \caption{Joint distribution of the dependent variables of study (\textit{number of parameters}, \textit{architecture type} and the \textit{dataset}) and the \textit{time of CPU usage}.}
    \label{RTP-full-img}
\end{figure}

In Figure \ref{RTP-full-img} we show the joint distribution between the \textit{number of parameters} and the \textit{time of CPU usage} for all different image datasets and architecture types.

\begin{figure}[h!]
    \centering
    \includegraphics[width=\linewidth]{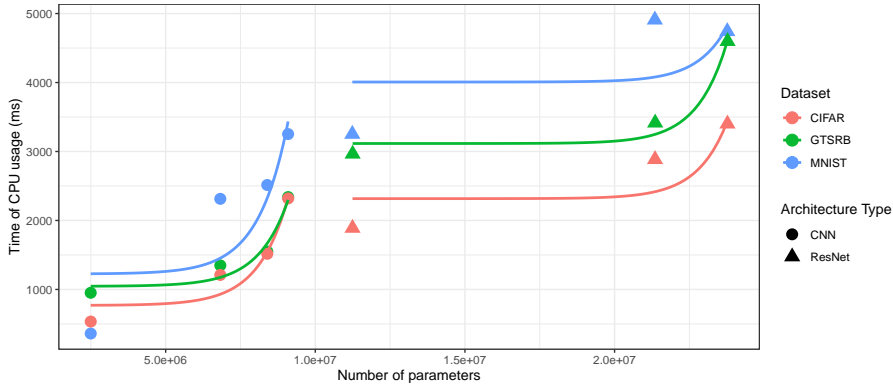}
    \caption{Joint distribution of the dependent variables of study (\textit{number of parameters}, \textit{architecture type} and the \textit{dataset}) and the \textit{time of CPU usage} of the C-CNNs.}
    \label{RTP-ccnns-img}
\end{figure}

\begin{figure}[h!]
    \centering
    \includegraphics[width=\linewidth]{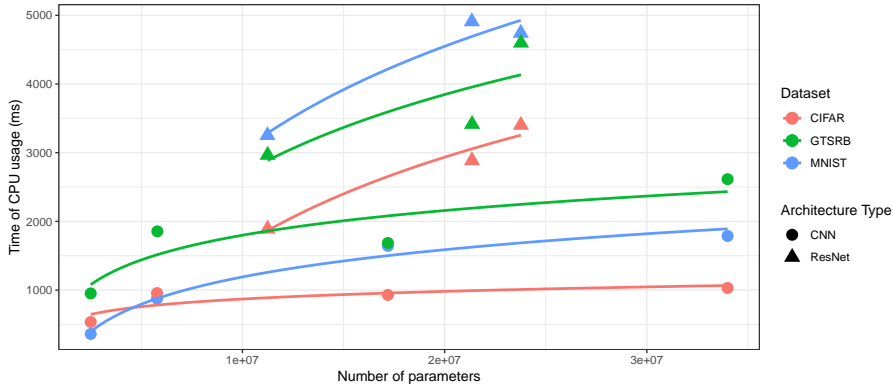}
    \caption{Joint distribution of the dependent variables of study (\textit{number of parameters}, \textit{architecture type} and the \textit{dataset}) and the \textit{time of CPU usage} of the FC-CNNs.}
    \label{RTP-fccnns-img}
\end{figure}

Regarding the \textit{architecture type}, Figure \ref{RTP-full-img} shows that for all ResNets there is a positive linear relation indicating that the more parameters of the models are used, the higher time cost in operation. However, the distribution of the \textit{time of CPU usage} shows a more complex shape for the CNNs (i.e. the CNN with higher time cost is not the one with more parameters). In the following we show the distribution of the \textit{time of CPU usage} for C-CNNs and FC-CNNs separately to depict more precise and meaningful insights. Figure \ref{RTP-ccnns-img} shows the distribution of the \textit{time of CPU usage} for C-CNNs and ResNets and Figure \ref{RTP-fccnns-img} shows the distribution of the \textit{time of CPU usage} for FC-CNNs and ResNets.

In Figure \ref{RTP-ccnns-img} it can be seen that the joint distribution of convolutional parameters and the \textit{time of CPU usage} is closer to a positive exponential form than to a linear one. This justifies that C-CNNs with a much smaller number of parameters than other FC-CNNs have a higher time cost in operation. This fact communicates that the convolution operation that happens in the first layers of all CNNs is significantly more time-consuming than the multiplications that happen in the FC layers.

Furthermore, in Figure \ref{RTP-fccnns-img} it can be seen that the joint distribution of fully-connected parameters and the \textit{time of CPU usage} is more similar to a logarithmic distribution rather than a linear one. This fact might be due to the hardware components of the operation devices optimizing its resources as they deal with bigger models and hence with more complex operations (i.e. parallel computations).

So far we have identified positive relations between the \textit{number of parameters} and the \textit{time of CPU usage} for both ResNets and CNNs in all the image datasets, even though they have different forms (i.e. linear, exponential, logarithmic). However, we notice that the datasets used also have an impact in the \textit{time of CPU usage} of the models, which might not be obvious if we assume that the \textit{time of CPU usage} of a model only depends on its specification (\textit{number of parameters} and \textit{architecture type}). Concretely, we identify that the differences in the \textit{time of CPU usage} within the three different image datasets studied are due to the differences in the \textit{number of parameters} that the \textit{dataset} involves. Specifically, there are two dataset-specific factors that have an effect in the \textit{number of parameters} and hence in the M:
\begin{itemize}
    \item 
{\it{(i)}} The number of channels of the images in the dataset affects the size of the first convolutional input layer in the models. For example, as MNIST contains gray-scale images, these are represented as one-channel tensors and hence the first convolutional layer has a depth of 1. In contrast, as GTSRB and CIFAR contain RGB images they require a model with a first convolutional layer of depth 3.\\

\item {\it{(ii)}} The number of classes in which we perform classification affects the number of output neurons in the last layer of the models and also the number of connections between the last two fully-connected layers.
For instance, MNIST models contain 10 output neurons corresponding to each of the existing digits and GTSRB models contain 43 neurons belonging to the different traffic sign classes in the dataset. If we suppose that the second last fully connected layer contained 128 neurons, MNIST models would have 128 * 10 connections + 10 neurons = 1,290 parameters and GTSRB models would have 128 * 43 connections + 43 neurons = 5,547 parameters in the last two layers.
\end{itemize}

In the following, we fit Model \ref{MModel} to describe the memory cost and test our hypotheses (see Section \ref{hypotheses}). Our hypotheses are based on testing what variables have a coefficient which fit is significantly different from zero. 

\begin{table}[h!]
\begin{center}
\caption{Summary of the \textit{time of CPU usage} Model fit in the image domain. The model reports \(R^2 = 91.54\) which defines a suitable goodness-of-fit.  \label{rtp-model-fit-img}}
\begin{tabular}{|c|l|l|l|c|}
    \hline
    \textbf{Variable} & \textbf{Coefficient} & \textbf{Coefficient fit} & \textbf{P-value} & \textbf{Significance}\\
    \hline
    P & $\beta_{P_1}$ & 2.79e-05 & 0.0761 & . \\
    AT-CNN & $\beta_{AT_{CNN_1}}$ & 7.45e+02 & 0.022 & * \\
    AT-ResNet & $\beta_{AT_{RN_1}}$ & 2.52e+03 & 2.62e-06 & *** \\
    D-MNIST & $\beta_{D_{MNIST_1}}$ & 6.65e+02 & 0.051 & . \\
    D-GTSRB & $\beta_{D_{GTSRB_1}}$ & 8.98e+02 & 0.011 & * \\
    \hline
\end{tabular}
\end{center}
\begin{tablenotes}
      \small
      \item The model is described as: $M = \beta_{P_1} P + \beta_{AT_1} AT + \beta_{D_1} D + \epsilon_1$
      \item Note that $\beta_{AT_1}$ and $\beta_{D_1}$ have a separate coefficient for each \textit{architecture type} and D respectively.
      \item Significance levels are: No, Very little '.', Little '*', Important '**', Very important '***' 
\end{tablenotes}

\end{table}

In Table \ref{rtp-model-fit-img} it can be seen that the variable that makes the most significant difference in the \textit{time of CPU usage} is the the \textit{architecture type}. According to the model fit, the CNNs used in this experiment imply a baseline time cost of 745ms and the ResNets of 2520ms. Also, each parameter adds an additional 2.79e-05ms. Additionally, experiments with GTSRB provide models that add a mean time cost of 665ms and in MNIST of 898ms.

For a more accurate estimate in the time cost that each parameter type adds, we fit the same Model \ref{MModel} for C-CNNs and FC-CNNs separately, which provides the results presented in Tables \ref{rtp-model-fit-ccnns} and \ref{rtp-model-fit-fccnns}. 

\begin{table}[h!]
\begin{center}
\caption{Summary of the \textit{time of CPU usage} model fit for ResNets and C-CNNs. The model reports \(R^2 = 96.54\) which defines a very suitable goodness-of-fit.  \label{rtp-model-fit-ccnns}}
\begin{tabular}{|c|l|l|l|c|}
    \hline
    \textbf{Variable} & \textbf{Coefficient} & \textbf{Coefficient fit} & \textbf{P-value} & \textbf{Significance}\\
    \hline
    P & $\beta_{P_1}$ & 1.52e-04 & 5.84e-05 & *** \\
    AT-CNN & $\beta_{AT_{CNN_1}}$ & 1.43e+02 & 0.63 & No \\
    AT-ResNet & $\beta_{AT_{RN_1}}$ & 1.84e+02 & 0.75 & No \\
    D-MNIST & $\beta_{D_{MNIST_1}}$ & 1.08e+03 & 0.01 & . \\
    D-GTSRB & $\beta_{D_{GTSRB_1}}$ & 4.87e+02 & 0.001 & **  \\
    \hline
\end{tabular}
\end{center}
\begin{tablenotes}
      \small
      \item Significance levels are: No, Very little '.', Little '*', Important '**', Very important '***' 
\end{tablenotes}
\end{table}

\begin{table}[h!]
\begin{center}
\caption{Summary of the \textit{time of CPU usage} model fit for ResNets and FC-CNNs. The model reports \(R^2 = 95.31\) which defines a very suitable goodness-of-fit.  \label{rtp-model-fit-fccnns}}
\begin{tabular}{|c|l|l|l|c|}
    \hline
    \textbf{Variable} & \textbf{Coefficient} & \textbf{Coefficient fit} & \textbf{P-value} & \textbf{Significance}\\
    \hline
    P & $\beta_{P_1}$ & 4.24e-05 & 0.003 & ** \\
    AT-CNN & $\beta_{AT_{CNN_1}}$ & 4.73e+01 & 0.878 & No \\
    AT-ResNet &  $\beta_{AT_{RN_1}}$  & 2.17e+03 & 1.25e-05 & *** \\
    D-MNIST & $\beta_{D_{MNIST_1}}$ & 8.49e+02 & 0.008 & . \\
    D-GTSRB & $\beta_{D_{GTSRB_1}}$  & 9.22e+02 & 0.013 & **  \\
    \hline
\end{tabular}
\end{center}
\begin{tablenotes}
      \small
      \item Significance levels are: No, Very little '.', Little '*', Important '**', Very important '***' 
\end{tablenotes}
\end{table}

In Tables \ref{rtp-model-fit-ccnns} and \ref{rtp-model-fit-fccnns} it can be seen that the estimated time costs for convolutional and fully-connected parameters are 1.52e-04 and 4.24e-05 respectively, indicating that a convolutional parameter implies a time cost approximately 3.5 times bigger than a fully-connected parameter. Finally, we summarize our findings in the distribution of the \textit{time of CPU usage} in image-based AI-enabled systems:

\begin{mdframed}
\noindent \textbf{Finding 1}: There exists a positive relation between the \textit{number of parameters} and the \textit{time of CPU usage}, meaning that an increase in the number of parameters of a CNN or a ResNet increases the time cost when operating with it. 

\noindent \textbf{Finding 2}: The \textit{time of CPU usage} increases close to exponentially when there is an increase in the number of convolutional parameters of a CNN.

\noindent \textbf{Finding 3}: The \textit{time of CPU usage} increases close to logarithmically when there is an increase in the number of fully-connected parameters of a CNN.

\noindent \textbf{Finding 4}: The image depth and number of classes in a dataset has an effect in the \textit{number of parameters} of the AI models that fit it, and hence in their \textit{time of CPU usage}.
\end{mdframed}

\subsubsection{Text-domain}
\begin{figure}[h!]
    \centering
    \includegraphics[width=\linewidth]{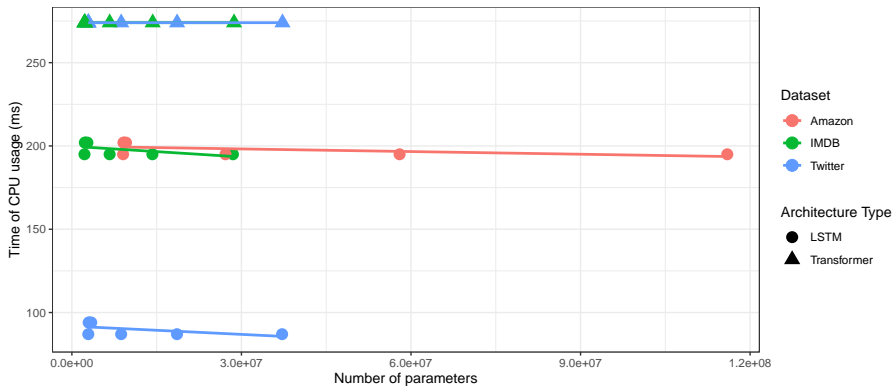}
    \caption{Joint distribution of the \textit{number of parameters}, \textit{architecture type}, \textit{dataset} and the \textit{time of CPU usage}. The figure does not show evidence of a postive nor negative relation between the \textit{time of CPU usage} and the \textit{number of parameters}.}
    \label{RTP-full-text}
\end{figure}

In Figure \ref{RTP-full-text} we show the joint distribution between the \textit{number of parameters} and the \textit{time of CPU usage} for all different text datasets and architecture types. 

\begin{figure}[h!]
    \centering
    \includegraphics[width=\linewidth]{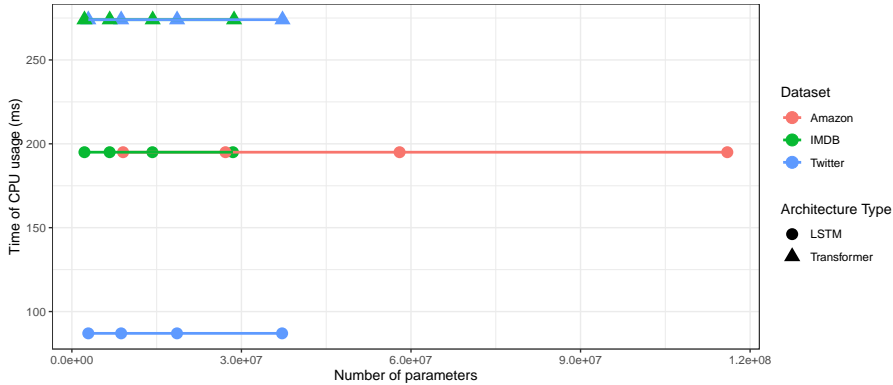}
    \caption{Joint distribution of the \textit{number of parameters}, \textit{architecture type}, \textit{dataset} and the \textit{time of CPU usage} of the EMB-extended models.}
    \label{RTP-emb-text}
\end{figure}

\begin{figure}[h!]
    \centering
    \includegraphics[width=\linewidth, height = 5cm]{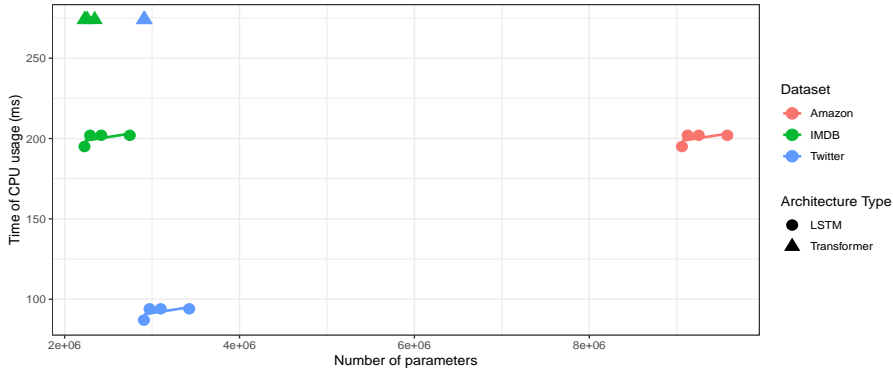}
    \caption{Joint distribution of the \textit{number of parameters}, \textit{architecture type}, \textit{dataset} and the \textit{time of CPU usage} of the NN-extended models.}
    \label{RTP-nn-text}
\end{figure}

Figure \ref{RTP-full-text} shows no evidence that increasing the number of parameters of a text model increases its time cost. However, it can be seen that Transformers have a higher time cost than LSTMs, and also that LSTMs in the Twitter dataset have a smaller \textit{time of CPU usage} than in the other two datasets. Note that the models have a different number of parameters depending on the dataset that they are trained on. This is due to the size of the embedding layer depending on the number of the unique words in each dataset, being 937,500 in the Amazon dataset, 234,375 in the IMDB dataset and 225,452 in the Sentiment140 dataset. Figure \ref{RTP-emb-text} shows the distribution of the \textit{time of CPU usage} for EMB-Transformers and EMB-LSTMs and Figure \ref{RTP-nn-text} shows the distribution of the \textit{time of CPU usage}for NN-Transformers and NN-LSTMs. 

In Figure \ref{RTP-emb-text} it can be seen that no matter what size the embedding layer has, the models need the same time to answer any query. As mentioned above, increasing the embedding parameters means increasing the size of a look-up table, and the fact that the time cost does not change is thanks to an efficient search algorithm on the look-up table which time complexity is independent of the size of the table.   Recall that the text models are a combination of the embedding layer (look-up table) and fully-connected layers (as is explained in Section \ref{modelling-section}). In this way, a fully-connected layer consists of doing matrix multiplication, and hence the bigger the size of the matrix the higher the time cost. However, for the embedding layer, no complex data structure operation is required but only a search on the look-up table. For this reason, the fact that increasing the \textit{number of parameters} does not increase the \textit{time of CPU usage} is thanks to the software used (i.e. Pytorch, Tensorflow) internally providing a search algorithm on the embedding layer that scales greatly with the size of the look-up table.

In Figure \ref{RTP-nn-text} it can be seen that the increase in the number of network parameters in the text models causes what looks like a logarithmic increase in the time cost. This is very similar to what we also observe in the FC-CNNs, and it makes sense because all FC-CNNs, NN-Transformers and NN-LSTMs have fully connected layers that work in the same way. In  Figure \ref{RTP-nn-text} it is difficult to see the latter for NN-Transformers and that is because we could not experiment with a wide range of number of network parameters since increasing them increases their memory cost drastically and we could not train them using our available computing environments (nor the one offered for free in Kaggle). That is, the memory cost of training such complex models is not accessible for researchers using standard computational environments but usually only available to big companies \cite{schwartz2019green}.

In the following, we fit the linear model \ref{MModel} to describe the \textit{time of CPU usage} in the text domain and test our hypotheses. Table \ref{rtp-model-fit-text} reports that increasing the number of parameters (both network and embedding parameters) does not have a significant effect over the \textit{time of CPU usage}. Also, it shows that LSTMs carry a baseline time cost of 117ms and Transformers of 247ms. Then, compared to the Sentiment140 dataset, IMDB dataset carries a baseline time cost of 53.8ms and the Amazon dataset of 82.4ms (e.g. fixing the value of the \textit{number of parameters}, a Transformer model in the amazon model has a time cost of 247ms + 82.4ms = 329.4ms for each query).

\begin{table}[h!]
\begin{center}
\caption{Summary of the model fit for the \textit{time of CPU usage} in the text domain. The model reports \(R^2 = 98.55\) which defines a very suitable goodness-of-fit.  \label{rtp-model-fit-text}}
\begin{tabular}{|c|l|l|l|c|}
    \hline
    \textbf{Variable} & \textbf{Coefficient} & \textbf{Coefficient fit} & \textbf{P-value} & \textbf{Significance}\\
    \hline
    P & $\beta_{P_1}$  & -6.28e-08 & 0.79 & No \\
    AT-LSTM & $\beta_{AT_{LSTM_1}}$  & 1.17e+02 & 5.39e-14 & *** \\
    AT-Transformer & $\beta_{AT_{Transformer_1}}$  & 2.47e+02 & 2e-16 & *** \\
    D-IMDB & $\beta_{D_{IMDB_1}}$ & 5.38e+01 & 7.01e-06 & *** \\
    D-Amazon & $\beta_{D_{Amazon_1}}$  & 8.24e+01 & 2.23e-06 & *** \\
    \hline
\end{tabular}
\end{center}
\begin{tablenotes}
      \small
      \item Significance levels are: No, Very little '.', Little '*', Important '**', Very important '***' 
\end{tablenotes}
\end{table}

\newpage
Finally, we summarize our findings in the distribution of the \textit{time of CPU usage} in text-based AI-enabled systems:

\begin{mdframed}
\noindent \textbf{Finding 5}: Contrary to what happens in the image domain, there is no evidence that there is a significant relation between the \textit{number of parameters} and the \textit{time of CPU usage}. Even though we conclude that the latter holds for embedding parameters, we believe that a more extensive analysis over the network parameters could provide evidence that increasing the \textit{number of parameters} also increases the memory and time cost. 

\noindent \textbf{Finding 6}: The number of unique words in a text dataset drastically conditions the number of parameters of the text models since it defines the size of the embedding layer. However, this does not have an impact on the \textit{time of CPU usage}.

\noindent \textbf{Finding 7}: Transformers have a baseline time cost that is higher than the one in LSTMs. 

\end{mdframed}

\subsection{RQ1.2: How is the storage weight of the AI models affected by their number of parameters, architecture type and dataset used?}
\subsubsection{Image domain}

\begin{figure}[h!]
    \centering
    \includegraphics[width=\linewidth]{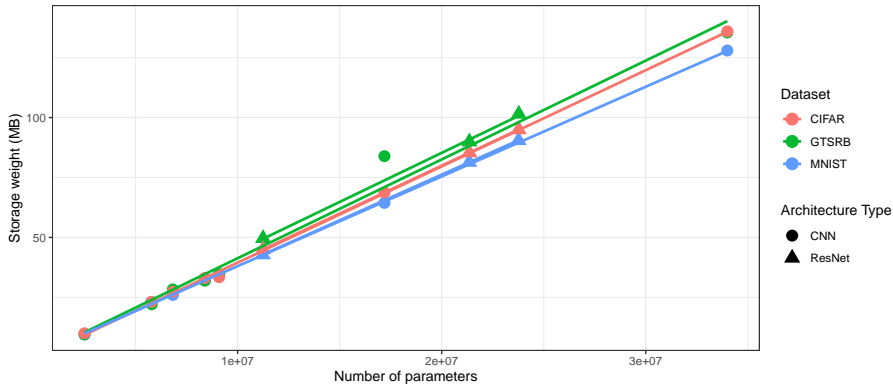}
    \caption{Joint distribution of the \textit{number of parameters}, \textit{architecture type}, \textit{dataset} and the \textit{storage weight}. The figure clearly shows a positive linear relationship between the \textit{number of parameters} and \textit{storage weight} for all architectures and datasets.}
    \label{SW-P-img}
\end{figure}

In the following we study the relation between the dependent variables of study (\textit{number of parameters}, \textit{architecture type}, \textit{dataset}) and the storage weight (S). In Figure \ref{SW-P-img} we show the joint distribution between the \textit{number of parameters} and \textit{storage weight} for both CNNs and ResNets models and all the image datasets.  

In Figure \ref{SW-P-img} it can be seen that there is a very clear and positive linear relationship between the \textit{number of parameters} and \textit{storage weight} for all architectures and datasets of study. In this case, the \textit{storage weight} distribution does not make a distinction between C-CNNs and FC-CNNs either. When fitting the distribution of \textit{storage weight} with respect to the three variables of the \textit{number of parameters}, \textit{architecture type} and the \textit{dataset} described in model \ref{SWModel}, we obtain the results shown in Table \ref{gtsrb-sw-fit}. The linear fit reports an \begin{math} R^2 \end{math} value of 0.99 which indicates that with the selected independent variables we are capturing all the variance in the target variable. Furthermore, we can see that only the \textit{number of parameters} is significantly causing variations in the distribution of \textit{storage weight}. Hence, we can provide a valuable estimated coefficient for the \textit{number of parameters} of value 3.96e-06. With this, we can make predictions on the storage weigh (S) of a model by only knowing its number of parameters: e.g. a CNN (or a ResNet) with 3,545,342 parameters will approximately weight 3,545,342 * 3.96e-06 = 14.04MB.

\begin{table}[h!]
\begin{center}
\caption{Results of the linear fit between the \textit{number of parameters} and the \textit{architecture type}, and \textit{storage weight}. \label{gtsrb-sw-fit}}
\begin{tabular}{|c|l|l|l|c|}
    \hline
    \textbf{Variable} & \textbf{Coefficient} & \textbf{Coefficient fit} & \textbf{P-value} & \textbf{Significance}\\
    \hline
    P & $\beta_{P_1}$ & 3.96e-06 & $<$2-16 & *** \\
    AT-ResNet & $\beta_{AT_{RN_2}}$ & 1.06e+00 & 0.5973& No \\
    AT-CNN & $\beta_{AT_{CNN_2}}$ & -1.97e-01 & 0.8929 & No \\
    D-GTSRB & $\beta_{D_{GTSRB_2}}$ & 2.99e+00 & 0.1646 & No \\
    D-MNIST & $\beta_{D_{MNIST_2}}$ & -2.46e+00 & 0.1250 & No \\
    \hline
\end{tabular}
\end{center}
\begin{tablenotes}
      \small
      \item The model is described as: $S = \beta_{P_2} P + \beta_{AT_2} AT + \beta_{D_2} D + \epsilon_2$
      \item Note that $\beta_{AT_2}$ and $\beta_{D_2}$ have a separate coefficient for each \textit{architecture type} and \textit{dataset} respectively.
      \item Significance levels are: No, Very little '.', Little '*', Important '**', Very important '***' 
\end{tablenotes}
\end{table}

We provide the following findings:

\begin{mdframed}
\noindent \textbf{Finding 8}: There exists a linear relationship between \textit{storage weight} and the \textit{number of parameters} described as $S = 3.96\mathrm{e}{-06} * P$.

\noindent \textbf{Finding 9}: No other variables significantly cause variations in the distribution of \textit{storage weight}.
\end{mdframed}

\subsubsection{Text-domain}
\begin{figure}[h!]
    \centering
    \includegraphics[width=\linewidth]{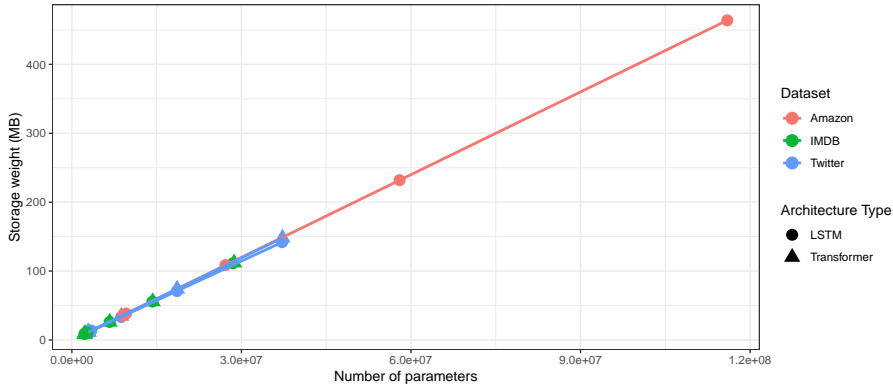}
    \caption{Joint distribution of the \textit{number of parameters}, \textit{architecture type}, \textit{dataset} and \textit{storage weight}. The figure clearly shows a positive linear relationship between the \textit{number of parameters} and \textit{storage weight} for all architectures and datasets.}
    \label{SW-full-text}
\end{figure}

In the following we study the relation between the dependent variables of the study (\textit{number of parameters}, \textit{architecture type}, \textit{dataset}) and \textit{storage weight} for text models. In Figure \ref{SW-full-text} we show the joint distribution between the \textit{number of parameters} and \textit{storage weight} for both Transformers and LSTMs and all the text datasets.  

As it happened with the image models, there exists a very clear and positive linear relation between the \textit{number of parameters} and \textit{storage weight}, and it can be seen for all text models and datasets. When fitting the linear model \ref{SWModel} to the distribution of \textit{storage weight} over the \textit{number of parameters} we obtain the results in Table \ref{text-sw-fit}.
\begin{table}[h!]
\begin{center}
\caption{Results of the linear fit between the \textit{number of parameters} and \textit{storage weight}. \label{text-sw-fit}}
\begin{tabular}{|c|l|l|l|c|}
    \hline
    \textbf{Variable} & \textbf{Coefficient} & \textbf{Coefficient fit} & \textbf{P-value} & \textbf{Significance}\\
    \hline
    P & $\beta_{P_2}$ & 3.97e-06 & 2e-16 & *** \\
    AT-LSTM & $\beta_{AT_{LSTM_2}}$  & 2.37e+00 & 0.7265 & No \\
    AT-Transformer & $\beta_{AT_{Transformer_2}}$  & 1.91+01 & 0.5918 & No \\
    D-IMDB & $\beta_{D_{IMDB_2}}$ & -6.14e-01 & 0.2739 & No \\
    D-Amazon & $\beta_{D_{Amazon_2}}$  & 3.80e+00 & 0.3561 & No \\
    \hline
\end{tabular}
\end{center}
\begin{tablenotes}
      \small
      \item Significance levels are: No, Very little '.', Little '*', Important '**', Very important '***' 
\end{tablenotes}
\end{table}

Again, we obtain a R2 of 0.99 indicating a quasi-perfect goodness-of-fit. The linear model reports that each parameters (be a network or embedding parameter) adds an additional 3.97e-06MB to the model's weight, which is a very similar result than the one obtained for image models, which was 3.96e-06MB.

In this way, we report an additional finding relating both domains.

\begin{mdframed}
\noindent \textbf{Finding 10}: In the text domain there also exists a clear and linear relationship between \textit{storage weight} and the \textit{number of parameters} described as $S = 3.97\mathrm{e}{-06} * P$.
\end{mdframed}

\subsection{RQ1.3: How is the accuracy of the AI models affected by their number of parameters, architecture type and dataset used?}

\subsubsection{Image-domain}
\begin{figure}[h!]
    \centering
    \includegraphics[width=\linewidth]{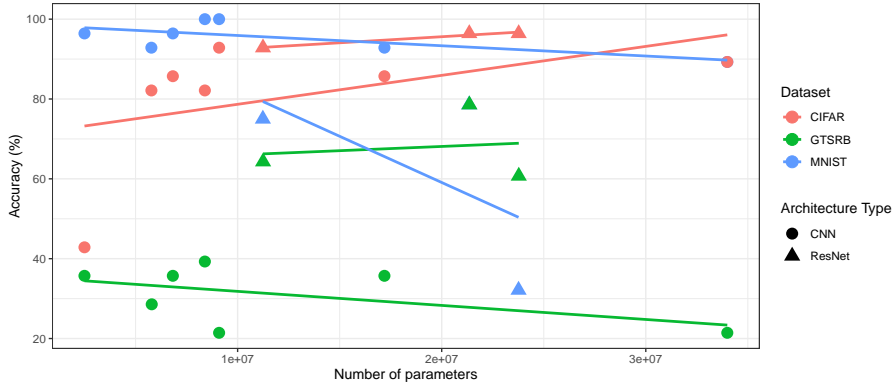}
    \caption{Joint distribution of the \textit{number of parameters}, \textit{architecture type}, \textit{dataset} and \textit{accuracy}. The figure shows a cloud of points which indicates that the distribution of \textit{accuracy} might be more complex than linear with respect to the independent variables.}
    \label{A-P-full-img}
\end{figure}

Finally, we study the joint distribution of the \textit{accuracy} and the different values of the \textit{number of parameters}, \textit{architecture type} and the \textit{dataset}. The overall distribution for the image-domain is shown in Figure \ref{A-P-full-img}.

\begin{figure}[h!]
    \centering
    \includegraphics[width=\linewidth]{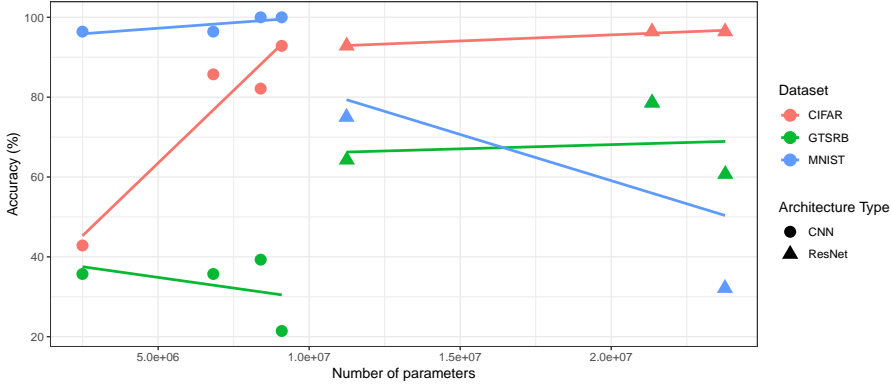}
    \caption{Joint distribution between the \textit{number of parameters}, \textit{dataset} and \textit{accuracy} for C-CNNs and in all the image datasets.}
    \label{A-P-C_CNNs}
\end{figure}

\begin{figure}[h!]
    \centering
    \includegraphics[width=\linewidth]{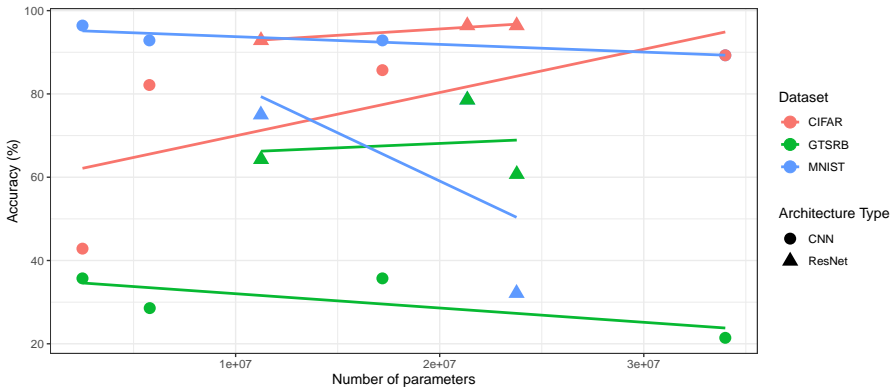}
    \caption{Joint distribution between the \textit{number of parameters}, \textit{dataset} and \textit{accuracy} for FC-CNNs and in all the image datasets.}
    \label{A-P-FC_CNNs}
\end{figure}

Inspecting Figure \ref{A-P-full-img} we observe that increasing the \textit{number of parameters} has a positive effect in the \textit{accuracy} in CIFAR but a negative effect in GTSRB and MNIST. To obtain more granular insights about the relation between the \textit{number of parameters}, \textit{architecture type}, \textit{dataset} and \textit{accuracy} we show the latter separately C-CNNs and FC-CNNs. In the following, Figures \ref{A-P-C_CNNs} and \ref{A-P-FC_CNNs} show the \textit{accuracy} achieved by C-CNNs and FC-CNNs.

A clearly delineated insight emerges from Figures \ref{A-P-C_CNNs} and \ref{A-P-FC_CNNs}: the dataset has a significant effect over the accuracy that the AI models achieve. Concretely, we identify MNIST as the easiest dataset for achieving the highest accuracy when using CNNs. However, ResNets appear to be a too complex architecture type for fitting the MNIST dataset. 

In addition, there is no common trend in terms of positive or negative relations between \textit{accuracy} and the \textit{number of parameters}, as the \textit{dataset} seems to have a very significant effect over the distribution of \textit{accuracy}. Specifically, the following results can be retrieved from the visualizations:
\begin{itemize}
\item {\it{(i)}} In MNIST, it seems that increasing the convolutional parameters of the CNNs helps the models extract better features from the images and hence obtain higher accuracy, but increasing the fully-connected parameters adds too much unnecessary complexity to the models and lowers their accuracy. Also, ResNets seem too complex models for MNIST as they obtain lower accuracy than basic CNNs.
\item {\it{(ii)}} In GTSRB, simpler is better, as the smallest models are the ones performing better. This can be possible if the distribution of the images can be easily separable among the 43 existing classes in GTSRB and hence the more complex models look for too sophisticated patterns for classifying the data. Also, ResNets perform much better than CNNs in operation because the data in GTSRB is very different from the one used in operation for measuring the accuracy (i.e. GTSRB contains traffic signs from Germany and operation data consists of traffic signs from Spain), and ResNets offer better generalization than CNNs.
\item {\it{(iii)}} CIFAR seems to be a too complex dataset for being overfitted and so the more complex the models, the better accuracy they achieve (i.e. the bigger the \textit{number of parameters}, the higher \textit{accuracy}, and ResNets perform better than CNNs).
\end{itemize}

\begin{mdframed}
\noindent \textbf{Finding 11}: The \textit{dataset} has a significant impact over \textit{accuracy}. Increasing the \textit{number of parameters} increases \textit{accuracy} only when there is enough data in the \textit{dataset}. If that is not the case, the distribution of \textit{accuracy} is non-linear and one needs to search for the sweet spot of the \textit{number of parameters} and the \textit{architecture type}.

\noindent \textbf{Finding 12}: Similar to the \textit{dataset}, the \textit{architecture type} also has a significant effect over \textit{accuracy}. Using more complex architectures (models that integrate more sophisticated pattern recognition operations, e.g. ResNets being more complex than CNNs) only has a positive effect over \textit{accuracy} when fitting large amounts of data that need such adjustments to be fit. 
In the other cases, using simpler models like CNNs works best, so it is always recommended to start with the simpler fits and add complexity incrementally only when needed. 
\end{mdframed}

\subsubsection{Text-domain}

In the following we show the distribution of \textit{accuracy} for text models, considering the EMB and NN models separately. 

\begin{figure}[h!]
    \centering
    \includegraphics[width=\linewidth]{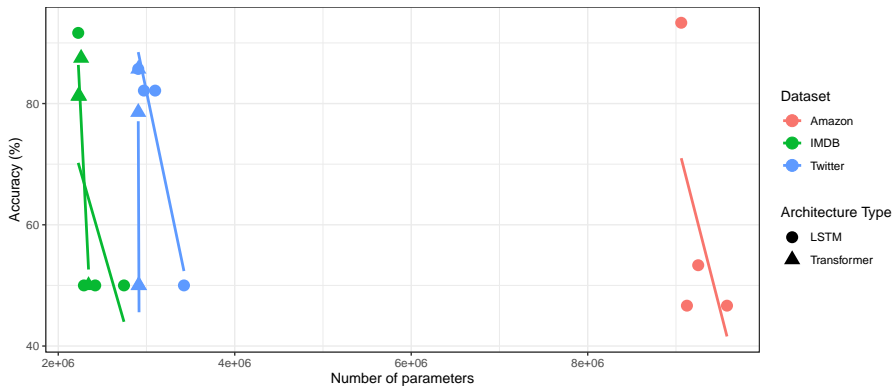}
    \caption{Joint distribution between the \textit{number of parameters}, \textit{dataset}, ATs and \textit{accuracy} for NN-models and all text datasets.}
    \label{A-nn-text}
\end{figure}

\begin{figure}[h!]
    \centering
    \includegraphics[width=\linewidth]{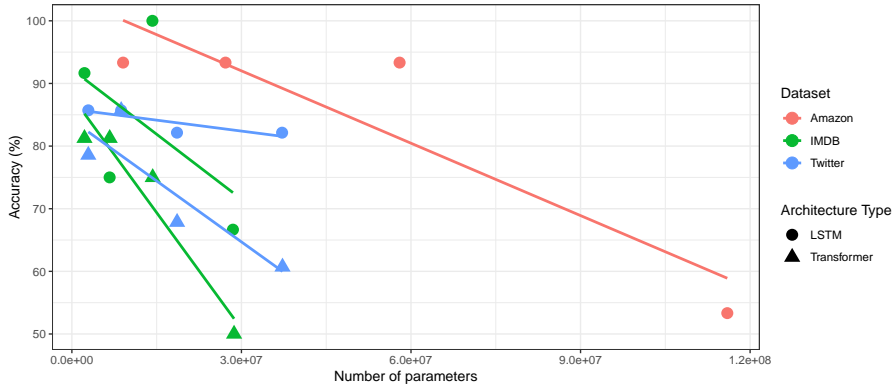}
    \caption{Joint distribution between the \textit{number of parameters}, \textit{dataset}, \textit{architecture type} and \textit{accuracy} for EMB-models and all text datasets.}
    \label{A-emb-text}
\end{figure}

In Figure \ref{A-nn-text}, we show the distribution of \textit{accuracy} for NN-Transformers and NN-LSTMs and in Figure \ref{A-emb-text} for EMB-Transformers and EMB-LSTMs. In both cases, we observe that for the selected datasets, increasing the \textit{number of parameters} (both network or embedding parameters) has a negative effect over \textit{accuracy}. The latter means that the simpler models are capable of learning the best language representations with the smallest embedding spaces and the way to relate and operate with them with the simplest non-linear functions (learnt by the fully-connected layers). In general, no dataset seems easier to solve than the others because the models achieve a similar accuracy on all of them. Also, generally the LSTMs achieve a higher accuracy than Transformers in the datasets of study.

\begin{mdframed}
\noindent \textbf{Finding 13}: For the three text datasets of study, the simpler models always achieve the best results. Hence it is recommended to start modelling the data in a small embedding space and with a small number of parameters.
\end{mdframed}

\section{Discussion}
In our work we have identified and depicted an underlying relationship between some design decisions of a set of AI models with respect to the overall performance of the mobile applications that integrate them.   In this section, we discuss (i) the subjective experience on the validity of profiling tools to analyze the accuracy-complexity trade-off of AI-enabled mobile applications and (ii) the implications of our results on both the AI engineering research and industry.

\subsection{Is profiling a suitable tool to systematically analyze the trade-off between accuracy and complexity of AI-enabled applications?}

In the following, we report our subjective empirical experience when collecting data using profilers to help determine whether profiling tools can become a standard approach for systematically reasoning about the performance of AI-enabled applications. In particular, we focus on the Android and Unity profilers, which are the ones used in this work.

\begin{figure}[h!]
    \centering
    \includegraphics[width=\linewidth]{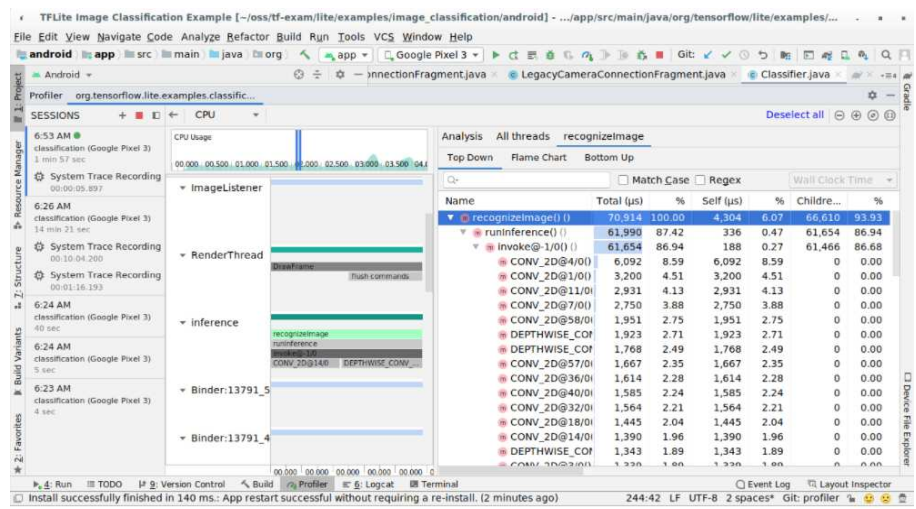}
    \caption{The Android Studio Profiler. It can be seen that each layer of the CNN is profiled separately, which is useful to get a better idea of which layers are the ones carrying more computations and resource usage.}
    \label{android_profiler}
\end{figure}

According to the Android Developers documentation, the Android Profiler tools provide real-time data to help one to understand how an app uses CPU, memory, network, and battery resources\footnote{\url{https://developer.android.com/studio/profile/android-profiler}}. Furthermore, the Android Profiler allows to create sessions, which allows to record data of different runs and compare it.  A drawback of the Android Profiler is that it is very sophisticated and that makes it quite difficult to use. It offers a lot of capabilities but are hard to configure. The good thing is that there are lots of official documentation which help to make the process easier. An example of usage of this tool consists of running an application, selecting which profiler module to view (CPU, memory, network, battery), selecting a time frame in the event timeline, and tracing the system calls. Regarding the latter, one can wrap a piece of code to be generated a as single system call in the profiler so the resources used are aggregated for all the operations that are wrapped (e.g. the \textit{predict} wrapper can aggregate all the convolutional and fully connected layer operations happening during a forward-pass of a CNN). All this capabilities are suitable for the developer experience on Android Studio. However, for this work we needed to statistically analyze the results with regression models and for that we needed an available dataset or data table that we can export and iterate externally. Although we have persistently tried to find a way to export the Android Profiler results to a readable data format (e.g. a CSV file) we have found no support for that, and hence we have not been able to do anything with the profiled results but seeing them displayed in our screens. The only supported export operation saves the profiled results in a \textit{.trace} file, which cannot be parsed into a format that can fit an statistical analysis with a software like R or Python. Figure \ref{android_profiler} shows the Android Profiler when computing an image class prediction with a CNN.

Regarding the Unity Profiler, the official documentation defines it as a tool that gathers and displays data on the performance of an application in areas such as the CPU, memory, renderer, and audio. It’s a useful tool to identify areas for performance improvement in the application, and iterate on those areas\footnote{\url{https://docs.unity3d.com/Manual/Profiler.html}}. As Unity is a graphical engine, the integrated Profiler provides modules like the illumination module, the physics, the UI, or the rendering ones. Then, just like the Android Profiler, it also provides the CPU and the memory modules, plus a GPU profiler. However, it lacks the network and battery modules. 

Compared to the Android Profiler, the one integrated in Unity is much easier to use, as one opens it and already sees the event timeline being updated in real-time. The user can click on a "Record" button to start tracing all system calls of the selected module. Then, it can stop recording and see the results in a sequential or hierarchical view. 

\begin{figure}[h!]
    \centering
    \includegraphics[width=\linewidth]{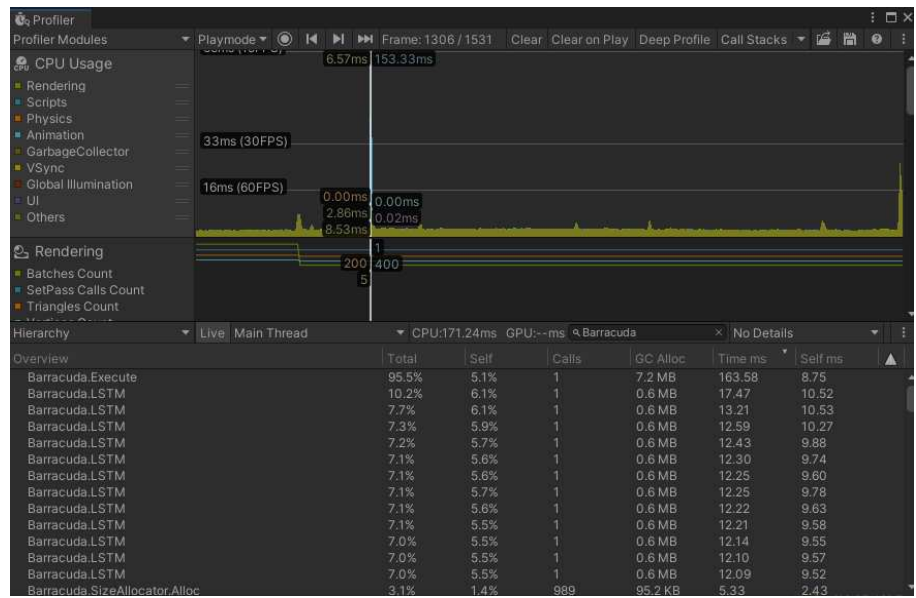}
    \caption{The Unity3D Profiler. As in the Android Studio Profiler, each of the different LSTM layers is profiled separately.}
    \label{unity_profiler}
\end{figure}

A great thing of the Unity profiler is that it can connect to an application run in an external device like a mobile phone. In this way, the profiler in the computer reports the resource usage of the external device, which is what we needed in this work, where we study the importance of the resource management in mobile devices. Figure \ref{unity_profiler} shows the Unity3D Profiler when computing a text sentiment prediction with a LSTM.

Finally, just like with the Android profiler, there is no integrated support in the Unity profiler to export the profiled data into an externally readable format. However, there is a tool in Unity named the \textit{Package Manager} which allows to import community and official packages to Unity, and we found a package named \textit{Profiler Analyzer} which allows the user to export the profiled data into a CSV file, which is what we needed for this work. 

For the availability to profile the performance of the mobile applications and to export the results into CSV files we mainly use Unity for analyzing the results of our experiments. However, we consider the Android profiler a very suitable tool for developers since it provides high-level insights of the performance of the applications. 

  Profilers were of critical relevance in this work, because without these tools we could not have any data at all to analyze. Hence, in this work, we argue that profiling tools are valid and useful to extract meaningful knowledge regarding the performance of AI-enabled applications. We consider that the two studied profilers are supported by rich documentation and are reasonably simple to use for practitioners, except for doing more complicated operations like exporting the results to specific formats. Finally, systematically comparing the capabilities of each profiler is out of the scope of this work, as it is to perform a more wide study using many other available and recent profiling tools (e.g. CodeCarbon \footnote{\url{https://codecarbon.io}}, Nvidia-smi \footnote{\url{https://developer.nvidia.com/nvidia-system-management-interface}}).

\subsection{Implications for AI engineering research}
Our work provides the AI engineering research community with an end-to-end experience on how to develop AI-based mobile applications, monitor their performance and interpret the profiling results. At the same time, we provide researchers with knowledge and awareness of the impact that the design decisions of AI models can have in the mobile applications' performance. Similar to other works like \cite{georgiou2022green}, we use Pytorch and Tensorflow as the DL frameworks for building a set of AI models with different specifications (in both their work and in ours we experiment with Transformers, CNNs and ResNets), and we also focus on monitoring the resource consumption of the AI models and describing statistically significant hypotheses. Compared to works like \cite{verdecchia2022data}, where the path to achieve greener AI is studied from a data-centric view, we work in a model-centric approach in which we focus on design decisions on the AI models (e.g. \textit{number of parameters} and \textit{architecture type}). Nevertheless, we also include the \textit{dataset} as a design decision that takes into account a data-based variable in the study of green AI.  Compared to the aforementioned works, we target mobile devices, which are environments with low computational resources, and we formulate linear models that can uncover significant relations between the models' configuration and the mobile applications' performance. 

Regarding the AI models that we experiment with (i.e. CNNs, ResNets, LSTMs, Transformers) we believe that the architecture and operations that these use are so different that it makes the \textit{architecture type} to play a key role in the results we obtain. In this sense, we consider that the impact of some design decisions (e.g. \textit{number of parameters} or \textit{dataset}) can be significantly different depending only on the \textit{architecture type}. 

Comparing Android Studio and Unity, which are the frameworks that support the development of AI-enabled mobile applications that we use in our work, we consider that Unity comes with a more user-friendly experience for developers. From feeding data to the AI models to profiling the applications' performance, Unity provides the same capabilities as Android Studio but requiring less coding of technical components. 

Finally, we provide the AI engineering research community with motivation to extend the degree in which this underlying relation can be characterised. One such example of the latter is the work by \textit{Verdecchia et al.} on studying what variables influence the performance of AI-enabled software from a data-centric approach. Other paths that we motivate are extending the set of design decisions (identifying new ones), including more metrics to study the impact to the applications' performance, using other frameworks that support the implementation of AI models in applications or working in edge devices other than mobile devices.

\subsection{Implications for AI engineering industry}
Our work provides the AI engineering industry with a detailed analysis of the impact of only working towards maximizing the accuracy of AI models without taking their complexity into account. In our work, we have shown that the more complex models are responsible of higher usage of resources and this correlates with more energy consumption. The trend of pursuing maximum accuracy models disregarding any other cost is more popular in the industry sector because companies want to take the maximum profit out of their AI models, and they usually have enough resources for doing so. However, with green AI becoming more and more popular and the need to reduce carbon footprint more important, the AI engineering industry has the urge to switch to greener AI-based solutions. We are living in a moment where the planet requires greener AI and the customers are gaining consciousness of the relevance of a company's compromise to work towards greener AI \cite{9585139}.

For these reasons, with our work we validate that a very suitable and feasible methodology for implementing and deploying AI models is to first make use of the simpler architectures (e.g. CNNs for image and LSTMs for text), and with a small number of parameters. From the results obtained, we have experienced that high accuracy results can be often obtained with models with such simple configurations, depending on the dataset used (which affects the quality and quantity of the data) and the problem to solve. For this reason, we argue that a suitable methodology for developing AI-enabled software is to start with the simplest solutions and add complexity only when needed. A valuable lesson learned that generalizes across the vision and language domains is that the distribution of the accuracy obtained with different models is difficult to frame with linear relationships, and hence we suggest tuning and evaluating the configurations of the AI models meticulously before considering more complex solutions. 

In particular, we suggest following the practices adopted in this paper for implementing the AI models:  \begin{itemize}
\item Have a finite set of models with substantially different scales to quickly gain insights on how the number of parameters is affecting the usage time of CPU and helping achieve higher accuracy (learned from Findings 1,2,3).

\item Keep the set of possible AI model scales small and have one for each specific architecture type of interest.

\item Starting with the biggest AI models, keep discarding the ones that do not significantly achieve higher accuracy. These will be the least valuable in terms of the accuracy-performance trade-off (learned from Findings 11,12,13).

\item Predict ahead the storage requirements of the model (following the formula uncovered in this manuscript) and see if it satisfies the application needs (learned from Findings 8,10).

\item Test its time of CPU usage and real-time performance in operation and see if it satisfies the application needs.
\end{itemize}

\section{Threats to validity}
As with any empirical study, there might be limitations to our study design. Next, we report a few mitigation actions taken during the design of this experiment.

\emph{Construct validity}: extent to which the theory constructs are correctly operated in the experiment. We build convolutional (i.e. baseline, ResNets), recurrent (i.e. LSTMs) and Transformer-based NNs in order to mitigate mono-operation bias.

\emph{Conclusion validity}: ability to draw correct conclusion between treatment and outcome, we define and fit linear models to approximate the relation between design decisions and accuracy and complexity-related metrics. Hence, we estimate the models in order to minimize the error when fitting the collected experimental data with linear relations. In this way, the results of our linear models report that these are able to capture almost all the variability of the performance metrics with respect to linear relations between the design decisions. With that and together with the statistical tests on the significance of each of the design decisions we validate our conclusions. Regarding the validity of the AI-enabled applications themselves, their quality depends on the experience of the developers involved since that can influence their complexity. For mitigating the latter we provide as simple implementations as possible by only implementing the required functionalities to the application and by following official tutorials for the development of these.

\emph{Internal validity}: extent to which statements can be made about the causal effects of treatments on outcome. A wide variety of NNs are developed and deployed using different technology stacks, mitigating that the results of our analysis of the overall performance of the applications are caused accidentally rather than by the variables of study themselves. Also, the error in the causal interpretation of our conclusions is tied to the estimation capabilities of our analysis models, which we report to be high.  

\emph{External validity}. Our results are tied to the datasets and technology used for experimentation, that is, on training the AI models with Pytorch and Tensorflow, exporting them with ONNX and TensorflowLite, and developing the mobile applications in Android Studio and Unity3D. For the datasets, the results are tied to the image datasets of GTSRB, MNIST and CIFAR, and the text datasets of Sentiment140, IMDB Movie Reviews and Amazon Reviews. All the aforementioned determines the experimental settings to which the results obtained that regard the accuracy and complexity could be generalized.

  Furthermore, we would like to remark that in this experimental work, we have followed the standards of empirical studies \footnote{\url{https://github.com/acmsigsoft/EmpiricalStandards/blob/master/docs/Experiments.md}}, where at the core, we manipulate independent variables, study dependent variables and apply each treatment independently to several experimental units. Our work contains the standard essential attributes of empirical studies, for instance: (i) we state formal hypotheses; (ii) we describe the dependent variables and justify how they are measured; (iii) we describe the independent variables and how they are manipulated or measured; (iv) we design an appropriate protocol for studying the stated research questions and hypotheses; and, (v) we discuss the interpretations of the results. Finally, our work also follows a set of other desirable attributes: (vi) we provide a replication package containing all the source code, analysis scripts and data used and collected; (vii) we show visualizations of the observed data distributions; and, (viii) we average our results across data of different sources and modalities.

\section{Conclusions and future work}

Our hypotheses have always been oriented to characterise an existing underlying relationship between the design decisions of AI models and the overall performance of the applications that integrate them. We have obtained results that prove that this relation exists and we have observed evidence of how adding more complexity to our AI models (more complex architectures or higher number of parameters) causes the models to require a higher usage of resources. We have tested this relationship in an environment in which we can derive statistically significant relations (i.e. a substantially big group of models with different configurations has been developed, variations on the design decisions have been iterated while keeping all other factors constant, and a total of 6 datasets in the vision and language domains have been used). Hence, we believe that our experimentation successfully targets the most popular AI models in the research community, and that provides meaningful knowledge to the vision and language-based communities. Finally, we have also provided experience on the end-to-end AI-enabled mobile applications development lifecycle, and have reported some of the issues that arise during such engineering process. During this lifecycle, we have also described our experience when using profiling tools to monitor the performance of AI-enabled mobile applications. With all this, we have provided important findings that arise from the results obtained. 

Additionally, we have provided practitioners from the AI engineering research and industry with awareness of how a set of design decisions on the AI models contributes to the overall performance of the AI-enabled applications. From this point, we motivate future work to characterise the relation that we have uncovered in more detail. For example, future work can {\it{(i)}} identify a bigger set of design decisions during the AI-enabled software engineering lifecycle; {\it{(ii)}} identify other green characteristics that can be monitored and studied as a function of the design decisions; {\it{(iii)}} extend the range of AI model architectures to study; {\it{(iv)}} generalise the edge devices to others than mobile devices; {\it{(v)}} extend the technology stacks used for experimentation; and {\it{(vi)}} provide experiences with other tools to monitor the performance of the AI-enabled applications.

\begin{acknowledgements}
This work has been partially supported by the GAISSA project (TED2021-130923B-I00, which is funded by MCIN/AEI/10.13039/501100011033 and by the European Union “NextGenerationEU”/PRTR); and, by the ``Beatriz Galindo'' Spanish Program (BEAGAL18/00064).
\end{acknowledgements}

\section*{Data Availability Statement}
We provide a public repository that can be found \href{https://github.com/roger-creus/Which-Design-Decisions-in-AI-enabled-MobileApplications-Contribute-to-Greener-AI}{here}. The repository contains {\it{(i)}} the source code to train all the AI models in the study, which includes the training datasets; {\it{(ii)}}  the source code of the AI-enabled mobile applications; {\it{(iii)}}  the evaluation datasets used to profile the metrics in the study during operation;  {\it{(iv)}}  the profiled datasets containing the values of the profiled metrics (i.e. accuracy, time of CPU usage, storage weight profiled during operation); and {\it{(v)}}  the source code to carry the statistical analysis of the profiled metrics.

\section*{Conflicts of Interest}
The authors declared that they have no conflict of interest.

%
%

\bibliographystyle{spphys}       
\bibliography{references.bib}   

\end{document}